\documentclass[final,3p,times]{elsarticle}

\usepackage{amsmath,amsthm,amssymb}

\usepackage{booktabs}
\usepackage[counterclockwise]{rotating}
\usepackage{multirow}
\usepackage{adjustbox} 

\usepackage{enumitem}

\usepackage{soul}

\usepackage[hidelinks, colorlinks=true,allcolors=blue]{hyperref}

\journal{Journal}

\usepackage{tikz,xcolor,hyperref}

\definecolor{lime}{HTML}{A6CE39}
\DeclareRobustCommand{\orcidicon}{%
	\begin{tikzpicture}
	\draw[lime, fill=lime] (0,0) 
	circle [radius=0.16] 
	node[white] {{\fontfamily{qag}\selectfont \tiny ID}};
	\draw[white, fill=white] (-0.0625,0.095) 
	circle [radius=0.007];
	\end{tikzpicture}
	\hspace{-2mm}
}

\foreach \x in {A, ..., Z}{%
	\expandafter\xdef\csname orcid\x\endcsname{\noexpand\href{https://orcid.org/\csname orcidauthor\x\endcsname}{\noexpand\orcidicon}}
}

\begin{document}

\begin{frontmatter}



\title{Telehealthcare and Telepathology in Pandemic: A Noninvasive, Low-Cost Micro-Invasive and Multimodal Real-Time Online Application for Early Diagnosis of COVID-19 Infection}


\author[inst1]{Abdullah Bin Shams\corref{cor1}\fnref{orcidA}}
\cortext[cor1]{Corresponding Author}
\ead{ab.shams@utoronto.ca}
\fntext[orcidA]{Joint first author, 0000-0003-0823-5333 \orcidA{}}
\affiliation[inst1]{organization={The Edward S. Rogers Sr. Department of Electrical and Computer Engineering, University of Toronto},
            addressline={10 King’s College Road}, 
            city={Toronto},
            postcode={M5S 3G4}, 
            state={Ontario},
            country={Canada}}
            
\author[inst2]{Md. Mohsin Sarker Raihan\fnref{orcidB}}
\ead{raihan1815505@stud.kuet.ac.bd}
\fntext[orcidB]{Joint first author, 0000-0002-0401-312X \orcidB{}}
\affiliation[inst2]{organization={Department of Biomedical Engineering, Khulna University of Engineering and Technology},
            city={Khulna},
            postcode={9203},
            country={Bangladesh}}

\author[inst3]{Md. Mohi Uddin Khan\fnref{orcidC}}
\ead{mohiuddin63@iut-dhaka.edu}
\fntext[orcidC]{ 0000-0002-1711-4237 \orcidC{}}
\affiliation[inst3]{organization={Department of Electrical and Electronic Engineering, Islamic University of Technology},
            addressline={Boardbazar}, 
            city={Gazipur},
            postcode={1704},
            country={Bangladesh}}
            
\author[inst5]{Ocean Monjur\fnref{orcidE}}
\ead{oceanmonjur@iut-dhaka.edu}
\fntext[orcidE]{ 0000-0003-1526-6808 \orcidE{}}
\affiliation[inst5]{organization={Department of Computer Science and Engineering, Islamic University of Technology},
            addressline={Boardbazar}, 
            city={Gazipur},
            postcode={1704},
            country={Bangladesh}}

\author[inst4]{Rahat Bin Preo\fnref{orcidD}}
\ead{rahatbinpreo@gmail.com}
\fntext[orcidD]{ 0000-0002-0026-5848 \orcidD{}}
\affiliation[inst4]{organization={Department of Electrical and Electronic Engineering, Bangladesh University of Business and Technology},
            addressline={Mirpur}, 
            city={Dhaka},
            postcode={1216},
            country={Bangladesh}}

\begin{abstract}
To contain the spread of the virus and stop the overcrowding of hospitalized patients, the coronavirus pandemic crippled healthcare facilities, mandating lockdowns and promoting remote work. As a result, telehealth has become increasingly popular for offering low-risk care to patients. However, the difficulty of preventing the next potential waves of infection has increased by constant virus mutation into new forms and a general lack of test kits, particularly in developing nations. In this research, a unique cloud-based application for the early identification of individuals who may have COVID-19 infection is proposed. The application provides five modes of diagnosis from possible symptoms (f1), cough sound (f2), specific blood biomarkers (f3), Raman spectral data of blood specimens (f4), and ECG signal paper-based image (f5). When a user selects an option and enters the information, the data is sent to the cloud server. The deployed machine learning (ML) and deep learning (DL) models classify the data in real time and inform the user of the likelihood of COVID-19 infection. Our deployed models can classify with an accuracy of 100\%, 99.80\%, 99.55\%, 95.65\%, and 77.59\% from f3, f4, f5, f2, and f1 respectively. Moreover, the sensitivity for f2, f3, and f4 is 100\%, which indicates the correct identification of COVID positive patients. This is significant in limiting the spread of the virus. Additionally, another ML model, as seen to offer 92\% accuracy serves to identify patients who, out of a large group of patients admitted to the hospital cohort, need immediate critical care support by estimating the mortality risk of patients from blood parameters. The instantaneous multimodal nature of our technique offers multiplex and accurate diagnostic methods, highlighting the effectiveness of telehealth as a simple, widely available, and low-cost diagnostic solution, even for future pandemics.
\end{abstract}


\end{frontmatter}

\section{Introduction}
\label{Introduction}
The present globe has been swept by the Corona Virus Disease since the first outbreak of Severe Acute Respiratory Syndrome Coronavirus 2 (SARS-COV-2) in 2019, commonly referred to as the COVID-19 pandemic. The virus has rapidly spread throughout all geographic regions transmitting the infectious airborne disease that has wreaked havoc on the healthcare industry and the well-being of the general public. COVID chronicles featured imposing long-lasting regional shutdowns multiple times, locking up people in quarantine and isolation centers, overburdening healthcare systems in most of the countries, significant disruptions of business flow along with supply chain, and most crucially, the deaths of a few millions of people worldwide to this date. 

Due to the vicious nature of rapid continual mutations of the spike-protein and RNA structure of the SARS-CoV-2 virus \cite{mutation1, mutation2} and the high rate of community transmission \cite{high_transmission}, some scholars fear that COVID-19 infection may not be permanently eradicated resulting in an everlasting endemic \cite{endemic} despite a massive vaccination program \cite{vaccine}. It is even impossible to anticipate the forthcoming variant of the virus due to the lack of pre-evidence of its mutating nature inside human bodies residing in a variety of socio-economical and geographical places around the world.

\begin{figure*}[ht]
\centering 
\resizebox{15cm}{!}{\includegraphics{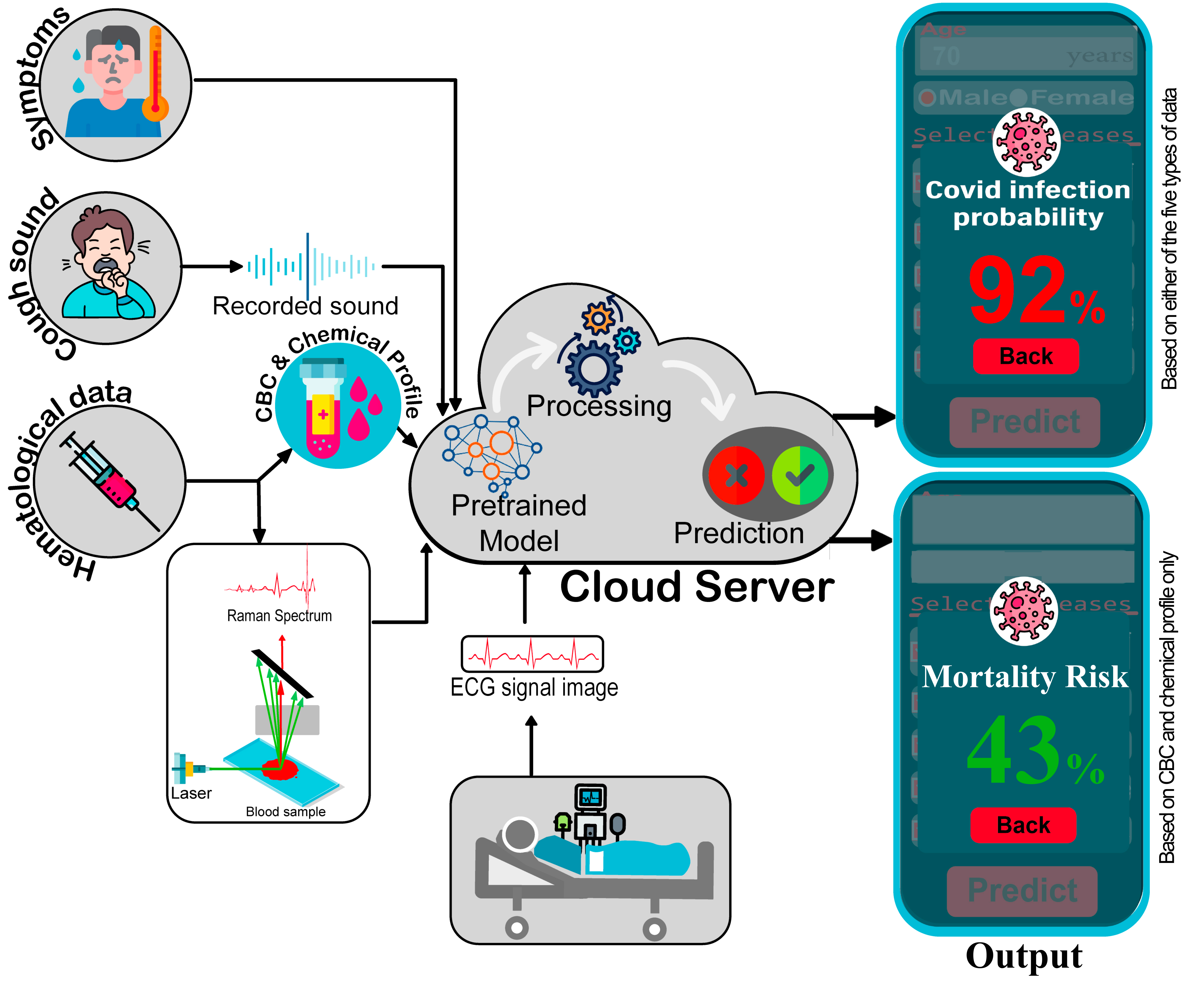}}
\caption{{A cloud-based framework for multimodal COVID-19 diagnosis and patient mortality risk prediction.}}
\label{fig1}
\end{figure*}

Furthermore, because the virus takes a long time to incubate \cite{incubation} inside the host body, it can spread quickly through infected individuals' respiratory droplets to healthy people via community transmission. Early detection of COVID-19 infection, tracking infected patients, contact tracing, timely treatment of the patient at risk and epidemiologic characterization is the imperative tasks to suppress this morbid situation quickly and to thwart the rapid transmission of the virus by making effective public health decisions for maintaining COVID-19 basic reproduction number \cite{blood1} far below from one. COVID testing on a large-scale employing point-of-care diagnostic procedures is an essential arsenal to achieve this goal. 

Vandenberg et al. \cite{test_limitation_1}, Mardian et al. \cite{test_limitation_2}, Scarpetta et al. \cite{test_limitation_3}, Pray et al. \cite{test_limitation_4} and Syal et al. \cite{test_limitation_5} carried out a comprehensive assessment on current state-of-the-art COVID-19 diagnostic tests, quality control and implementation issues, diagnostic tool limitations such as in low-resource settings, and potential solutions to these kinds of shortcomings for future development of diagnostic procedures. A number of molecular tests for viral presence detection in respiratory droplets like nasopharyngeal or oropharyngeal swabs and immunoassays for virus-specific antibody or antigen detection in blood serum have been proposed by healthcare manufacturers since the onset of the pandemic to satisfy various concurrent clinical priorities. Amongst all, the FDA and WHO have approved the real-time polymerase chain reaction (RT-PCR) molecular test and the IgG/IgM Rapid Antigen/Antibody test as the gold standard confirmatory tests for COVID-19 diagnosis. Albeit, emerging and impoverished countries having low resource settings are facing a growing number of difficulties incorporating these tests in a mass point-of-care context due to logistics, capacity, and fiscal constraints. In RT-PCR test, the sample collection procedure creates discomfort among the patients and has a high chance of spreading the infection as the procedure is pretty risky due to the interaction between the healthcare professional and the patients. The SARS-CoV-2 present in collected nasal swab specimens from symptomatic patients goes through a virus culture process for nucleic acid amplification that requires costly chemical reagents and controlled environment created inside a sophisticated and expensive PCR machine. Then the direct presence of virus protein in the nasal swab is detected via chemical reaction using the PCR machine. The turnaround time is approximately 4-5 hours. The test accuracy varies depending on the viral load as the viral load reaches its zenith during the first week of the disease, and gradually declines with the progress of the illness and thus elicits false detections of about 38\%-67\%. Direct detection of viral presence in the samples may lead to erroneous false-negative results due to a technical error, as well as, false-positive results due to cross-contamination caused by poor preservation and time delay between sample collection and test. For these reasons, it is still debatable whether these tests truly fulfill the prerequisites of a point-of-care diagnostic method. The point-of-care method \cite{pointcare1, pointcare2} should be a fast, inexpensive, and robust diagnostic procedure capable of being conducted in a non-laboratory environment, which is a dire requirement during the COVID-19 outbreak. Many countries are still lacking in and thriving to ensure skilled personnel, sufficient reagent and PCR machine supply to accomplish these intricating tasks with minimal error to conduct bulk testing with expensive tools like RT-PCR.

Machine Learning (ML), Deep Learning (DL) and Artificial Intelligence (AI) have seen a plethora of wide-ranging novel applications since the outbreak of the pandemic commenced, such as real-time alerts to users about false information during online health inquiries over a web search engine \cite{SEMiNExt}, COVID-19 drug discovery \cite{drug}, and vaccine development \cite{vacdev}. The most popular work focused on identifying positive COVID-19 patients from the X-Ray images with satisfying outcomes. Unfortunately, recent extensive studies demonstrated the lacking potential of this technique for clinical use, ascribing methodological flaws or underlying biases.

Jiao et al. \cite{xray1} performed a detailed study on COVID-19 disease severity (critical vs non-critical) and disease progression prediction (prognostics) employing AI-based Chest X-ray (CXR) and twelve types of clinical data combined analysis method experimented in multiple hospitals located in Philadelphia, PA, USA for training-validation-internal testing of the proposed model and also performed an external test in multiple hospitals situated in Providence, RI, USA for which the proprietary dataset had been collected during March till July of 2020. The authors pointed out that Chest CT images are three-dimensional requiring high computational resources which inspired them to develop resource optimized two-dimensional CXR prognostic AI model. Resized ($512\times512$ px) and normalized $(0-1)$ images had been segmented using U-net architecture followed by feature extraction via VGG-11 visual geometry group architecture employing softmax classifier, adam optimizer with 0.0005 learning rate and negative log-likelihood loss. EfficientNet-B0 was used to combine masks of processed and segmented images followed by passing through four-layer deep CNN for image-based severity prediction. Clinical features had been passed through three-layer DCNN for clinical-data-based severity prediction and finally weighted (image + clinical) severity was predicted. AI-based CXR combined with clinical data analysis for severity prediction achieved a ROC-AUC score of 0.821 (95\% Confidence Interval - CI) to 0.846 $(p<0.0001)$ in the case of internal testing and for external testing, the score is (slightly less) 0.731 to 0.792 $(p<0.0001)$. Image-based DCNN layer outputs and clinical features had been passed through the Survival Forest algorithm individually followed by combination with CXR severity scores for combined progression prediction. Deep-learning feature combined with clinical data for progression prediction showed concordance index (C-index) from 0.769 to 0.805 $(p<0.0001)$ in case of internal testing and for external testing, the score is (slightly less) 0.707 to 0.752 $(p<0.0001)$. CXR and clinical data combined AI model showed higher performance than clinical data combined with radiologist’s manually derived severity score model: (C-index 0.805 vs 0.781; $p=0.0002$) in case of internal testing and for external testing, the score is slightly less (C-index 0.752 vs 0.715; $p<0.0001$). Kaplan-Meier curves and time-dependent ROC-AUCs for progression risk prediction are also depicted in the corresponding article.

New York University Langone Health based research experiment conducted on admitted patients (19,957 CXR images collected during March 3 - May 13, 2020) by Shamout et al. \cite{xray2} developed and deployed an automated AI model for COVID-19 patient deterioration (intubation, admission to the ICU and in-hospital mortality) risk score prediction using a combination of Deep CNN from pre-processed CXR images and Gradient Boosting prediction from routine clinical variables (fifty-eight input features in total). Clinical variables (vital signs and lab-test results), patient characteristics and adverse events are tabulated with distribution in the corresponding article. The deterioration (in a timeframe within 24, 48, 72, 96 hours) prognostic model achieved a ROC-AUC score of 0.786 (95\% CI: 0.745–0.830) during the test. Their proposed Globally-Aware Multiple Instance Classifier (GMIC) model also computed the saliency maps on the CXR images marking the adversely affected region (airspace opacities and consolidation) followed by segmentation of these Region of Interest (ROI) patches. GMIC Deterioration Risk Curve model plots the time evolution of lung condition deterioration. Detailed architecture of GMIC model comprising Global, Local and Fusion module is portrayed in the corresponding article. Detailed results of GBM, GMIC and ensemble model of the corresponding article indicate that the GBM+GMIC ensemble model performs best and outperforms two radiologists’ (with seventeen and three years of experience) manual deterioration score prediction. DRC curves plotted using the GMIC-DRC model clearly distinguishes between patients with and without adverse events.

Despite all the success stories, DeGrave et al. challenged the robustness and trustworthiness of the AI-based CXR (also applicable to Chest CT images) analysis models through the University of Washington based Su-In Lee’s laboratory experiment \cite{xray3} due to unsatisfactory performance of the models tested at newer hospitals indicating the AI-based models might be learning counterfeit shortcuts instead of medical pathology at the institutions they had initially been trained on. For the experiment, the authors combined four \cite{xray4, xray5, xray6, xray7} online available datasets into two datasets; Dataset-1 consisted of COVID positive images from one online available dataset and COVID negative images from another one. Then Dataset-2 had been prepared incorporating both positive and negative images with equal proportion from remaining each of the two online available datasets. After detailed image preprocessing, datasets had been trained with a series of ten models of different architectures; also examined using multiple secondary models viz. COVID-Net network and CV19-Net. Testing the models on internal dataset and cross-dataset revealed that performance deteriorated to 50\% in the case of the cross-dataset test indicating reduced ability to perform well on generalized datasets which computer scientists call the Domain Shift Problem. The authors furthermore carried out a more complex investigation to find out which `shortcuts’ the AI models are learning. Investigating all the images via plotting saliency maps revealed that it pointed out irrelevant texts, arrows and image corners on CXR images as important markers along with lung fields. Hence, they transformed COVID +ve images as if they were -ve and vice versa by altering arrows, changing lung-field opacity and radiopacity at the image borders by using the Generative Adversarial Networks (CycleGANs). Testing these images via those previously trained AI models now started giving predictions on the opposite class. Therefore, they concluded that the proposed CXR analysis models so forth might be learning shortcuts by considering text-markers, arrow styles, etc. on the CXR images as important pixels. The authors also quoted another research article which showed an analysis of the CXR image with only image borders after removal of the COVID pathologies from all the images retained the model's high accuracy indicating the wicked learning nature of the CXR-AI models instead of pathological learning.

Alternatively, COVID-19 diagnosis from pathological parameters holds a great promise, as they are directly linked to the change in bodily function. Coronavirus infects people with mild to serious illnesses. The damage to multiple organs is associated with a massive spike in inflammatory markers \cite{cytokine1, cytokine2} and changes in other hematological parameters \cite{Hematological1, Hematological2}. This is followed by the development of symptoms such as fever, loss of taste or smell, shortness of breath, sore throat, cough, waist pain, etc. Point-of-care diagnosis based on the amalgamation of biosensors and telemedicine \cite{pointcare3, pointcare4, pointcare5} can emerge as a viable option in this situation. Biosensor devices diagnose medical conditions by detecting biomolecules, in other words, biochemical patterns in the body. As a result, the test reports are precise. Biosensing demands little to no processing of the sample. Artificial Intelligence and Biosensor dependent point-of-care diagnostic methods have considerable potential to generate fast and high throughputs at a low cost and put a great demand in the realm of telehealthcare. Telehealth is the new paradigm in health care, integrating digital information and communication technologies, to remotely deliver medical services. Conventionally, it was limited to ambulatory care such as patients in rural areas, older adults, and people with limited mobility. The telehealthcare concept promises to connect people to in-demand specialists and provide affordable quality healthcare services.  A recent study \cite{telemedicine1} shows that around two-thirds of patient prefers the convenience of telemedicine and the need for rapid access to their health records by their practitioners. Following the COVID-19 outbreak, a massive increase of 154\% in telehealth users has been recorded \cite{telemedicine2}. This can solely be attributed to a consensus on providing safe healthcare amidst the pandemic and reducing the risk of exposure to the virus. The rising popularity of telehealth brings the challenge of handling an enormous amount of data. To make the system more efficient Artificial Intelligence (AI) capabilities \cite{telemedicine_AI} can be integrated to detect patterns, point out potential issues, and aid doctors to act quickly in case of an emergency. 

On top of that, the prognosis of critically ill COVID-19 patients has shown a crude fatality rate ranging from 25\% to 50\% \cite{mortality1}. Several retrospective studies have analyzed the risk factors that adversely affect the health of COVID-19 infected patients including comorbidities like obesity, chronic cardiac and pulmonary conditions, hypertension, diabetes, chronic kidney disease, renal replacement therapy, cancer, development of ARDS, and dysfunction of organs during admission at hospital cohort \cite{mortality2, mortality3}. A massive inflammatory response and cytokine storm can be seen in the bloodstream of an infected patient which resembles gradual degradation of the health condition of the infected patient \cite{mortality4}. Therefore, a blood sample analysis along with symptoms of an infected patient with worsening lymphopenia, neutrophilia, and troponin elevation is a notable researched study to benchmark the mortality risk of critically ill patients \cite{mortality2, mortality3}. At the height of the global pandemic, it has been crucial to coequally distribute medical healthcare and facilities to every infected person. Therefore, taking emergency decisions and managing the limited medical health care facility for everyone has been a huge clinical challenge. Along with the shortage crisis of ICU beds, the lack of properly trained personnel to treat a huge influx of critically ill patients have been concerning to everyone \cite{mortality5}. A machine learning model to assist prediction of the mortality of patients, which health workers are often unable to provide at a mass rate in case of emergency during this pandemic is an essential tool. 

\vspace{3mm}

\noindent Based on these persuasion and clinical identifiers mentioned so forth, 

\vspace{2mm}

\begin{itemize}[noitemsep, topsep=0pt, leftmargin=*]

\item This research work sought to implement point-of-care diagnosis using five different modes of diagnostic procedures. The study proposes and experimentally demonstrates a novel online framework to identify COVID-19 infection from the post-infection patient symptoms, cough sound, hematological parameters from complete blood count and chemical analysis, blood specimen Raman spectroscopy image, and ECG signal image. The concept is illustrated in Figure-\ref{fig1}.

\item All the attributes associated with symptoms, cough sound, and blood parameters are optimized and only the most important features for accurate predictions are highlighted. This is significant to reduce the cost of COVID-19 detection from routine blood tests, which are available in most clinical and rural hospitalized settings. Therefore, our technique can easily be integrated locally with available medical infrastructure in developing and underdeveloped countries, where there is a scarcity of COVID-19 PCR test kits \cite{kit_shortage}, to provide an alternate feasible low-cost method of diagnosis.

\item The multimodal approach offers multiplex diagnostic methods to better classify possible infections.

\item Another aim of this study is to build an application-based tool that can assist during this crisis in predicting the mortality rate of a critically infected person from their blood sample analysis. The Artificial Intelligence-based model used in the application is capable of predicting mortality with very few parameters, thus decreasing the overall number of blood test reports required to identify a critically ill patient. This can overall reduce the cost and time required for both patients and health care workers during the crisis. Additionally, the identification of critically infected patients can assist in better distribution of the limited mechanical ventilation and ICU beds.

\item The online, scalable and real-time nature of our detection platform can be accessed by both non-ambulatory and hospitalized patients likewise. Predominantly, it can be used for rapid mass testing and mortality risk prediction both at low resource setting localities as well as overpopulated places like airports, train stations, etc. This can provide an additional fast and immediate screening method to limit the spread of the virus. 

\item The developed platform can be accessed through an online application, available for free from the following website: Supplementary Material - \ref{sup_mat}(a)

\end{itemize}

\noindent Remaining of the article is organized as follows: Section-\ref{Background Study} gives a theoretical review on the clinical manifestations of COVID-19 and a comprehensive review on the recent proposed approaches of COVID-19 diagnosis using artificial intelligence under multiple modalities. Section-\ref{Methodology} outlines the overall methodologies adopted during the research work including short description on dataset overview, data pre-processing steps, machine learning and deep learning algorithms used in the study, and the working principle of the proposed online application. Detailed results are discussed in Section-\ref{Result} and the article is concluded with Section-\ref{Conclusion}.

\section{Background Study} 
\label{Background Study}
This section presents a conspectus of the contemporary studies that used diagnostic modalities similar to this study, either to support the theoretical foundation or to support the AI-based diagnostics of COVID-19.

\subsection{Pathophysiology, etiology and theoretical review on clinical manifestations of COVID-19}
\label{Pathophysiology}

COVID-19 infection causes an infected person's immune system to be drastically impaired \cite{immune_system1, immune_system2, immune_system3, immune_system4}, resulting in progressively more severe health complications. The human lungs have the highest concentration of Angiotensin-Converting Enzyme-2 (ACE-2) that binds to the SARS-CoV-2 viral spike protein in type-2 alveolar cells. Air-duct and alveolar inner-surface infection caused by the virus initially triggers the innate immune system of the infected individual. Macrophages, dendritic cells, and natural killer cells (NK cells) combat the virus through strong cytolytic functions and try to avoid further inflammation \cite{immune_system3}. Just as the first approach fails, the virus gradually replicates inside the host alveolar cells and damages them via a cytokine storm loop, culminating in COVID-19 symptoms and progressive respiratory malfunctions leading to the need for mechanical ventilation and ICU support. Sore throat, forced cough, fever, fatigue, headache, nausea, vomiting, stomach pain, anorexia, and gastrointestinal hemorrhage are some of the early symptoms. Some people also experience rash and edema on their skin.  The virus may further evoke pneumonitis, pulmonary edema, dyspnea, hypoxemia, and acute respiratory distress syndrome (ARDS). When all other treatments are ineffective, the patient will succumb to death either from stroke or suffocation or multiorgan failures, such as kidney failure due to proteinuria-hematuria or liver failure due to elevated bilirubin-AST-ALT-LDH, as induced by systemic inflammation according to Diamond et al \cite{immune_system3}. Moreover, owing to distinct changes in airway epithelium and immune cells between pediatric and adult patients, Yoshida et al. \cite{immune_system2} observed that the systemic immune response in children is comparatively different from adults. 

The cardiac system and hence the immune system along with the patient's bloodstream possess specific traits and clinical changes \cite{Hematological1, Hematological2} that can be detected from certain pathological analyses. Toledo et al. \cite{Hematological3} have published a comprehensive clinical analysis of the parametric effects on complete blood count and hematochemical profile anomalies in patients with antagonistic COVID-19 infection. The physical phenotype (deformation, stiffness-change, changes to cell count, protein damage, cytoskeletal alterations, etc.) and mechanical properties of five common cell types (erythrocytes, lymphocytes, monocytes, neutrophils, and eosinophils) were found to be altered according to Kubankova et al. \cite{Hematological4}, illustrating the semi-permanent impact of COVID-19 on the patient-body.

Taking these considerations into account, the authors of this study used both hematological-hematochemical and computer vision evaluations of collected blood specimens in this proposed research work. Blood-cell count and chemical profile of blood-serum altercations in the COVID-19 infected patient's bloodstream can be measured with routine pathological blood tests like Complete Blood Count (CBC), hematochemical profile analysis, etc. One way to detect phenotypic changes in blood cells and the presence or changes in concentration of certain hematochemicals (e.g. immunoglobulins like IgG-IgA-IgM, albumin, fibrinogen, phospholipids, glycerolipids, transferring, cholesterol etc.) in the blood serum due to post-infection immunological response can be Raman spectroscopic raw data or image analysis obtained from either blood smear taken atop a glass microscope slide or blood-specimen concealed within cryopreservation tube for Raman spectroscopic analysis. Interestingly, whether only the separated serum after centrifuging the blood specimen should be used or the whole blood specimen containng all blood-cells should be used for COVID-19 diagnosis via Raman spectroscopy should be investigated further. The relevant conducted studies so forth have used only centrifuged blood serum, however.

Raman spectroscopy uses the inductile scattering of light to identify the vibrational states (phonons) of molecules. The biochemical composition of the sample is represented by the characteristic peaks of stokes or anti-stokes lines in the Raman spectrum, dubbed as the `fingerprints’ which are unique for each constituent of the specimen. Moreover, the capability of the Raman spectrum to gather and interpret spectral data from multiple positions on a sample to provide a statistically significant characterization of the sample’s polymorphism and multicomponent configuration leads to the detection of various biomarkers simultaneously \cite{raman1}. The intricate changes in the spectrum caused by a virus or other organisms in the human body can be well scrutinized using deep learning techniques due to its ability to adapt to transitions and the sensation of invisible trends \cite{raman2, raman3}. Bio-marking through the Raman scattering and analysis of the spectral images using deep learning algorithms like Convolutional Neural Network can serve as a portable, non-invasive, quick, contact-free and point-of-care diagnostic method \cite{raman4, raman5}.

COVID-19 infection also affects the usual cardiac rhythm of a patient. Brit Long et al. started a rigorous review on 200 research articles and case studies initially and published a review article \cite{ecg_reveiw1} on 80 COVID-19 induced cardiac emergency-medicine relevant articles and case studies selected by consensus. Authors narrated the underlying pathophysiology of various complications with relevant 12-lead ECG tracings of each kind and reported that COVID-19 can eventually affect the cardiovascular system of 90\% of critically ill patients causing a variety of cardiac complications which lead to ECG abnormalities. Studies conducted in France, Italy and United States revealed an increase (more than 52\%) in the number of Cardiac arrest dysrhythmias during the COVID-19 pandemic compared to pre-pandemic years. ECG manifestations along with probable causes and observed abnormalities are tabulated in Table-\ref{table1}.

\begin{table*}[htp]
\caption{ECG manifestations along with probable causes and observed abnormalities as surveyed by Brit Long et al. \cite{ecg_reveiw1}}
\centering
\begin{adjustbox}{width=\textwidth}
{\begin{tabular}{lll} \toprule

{\bf{Probable underlying causes}} & {\bf{ECG manifestations of COVID-19}} & {\bf{Cardiac abnormalities}}
\\ \midrule

hypercoagulability, plaque rupture, & changes in axis and intervals & sinus tachycardia (most frequent), \\

endothelial/myocardial injury, sepsis, & such as pathologic Q waves, QRS & ventricular arrhythmia (ventricular \\

hypoxic injury, coronary spasm, & complex axis deviation, QT & tachycardia/fibrillation, etc.), \\

cytokine storm, electrolyte & prolongation, ST-elevation and & supraventricular tachycardia (atrial flutter/ \\

abnormalities, cardiac interstitial & plateau, T wave inversion, & fibrillation, etc.), several bradycardias, \\

inflammatory infiltration, vascular & Brugada-like pattern, etc. & cardiac tamponade, myocarditis, torsade-\\

inflammation and necrosis, &  & de-pointes, ventricular block, LBBB, RBBB,\\

microthrombi, etc. & &  pulmonary embolism, cardiac arrest, etc. \\

\bottomrule
\end{tabular}}
\end{adjustbox}
\label{table1}
\end{table*}

Inspired by Brit Long et al., Alsagaff et al. \cite{ecg_reveiw2} further analyzed ECG data of COVID-19 patients admitted to hospitals collected from seven studies during the entire year of 2020 and tabulated the parametric values of COVID-19 manifested ECG wave segments and patient outcomes. From echocardiogram analysis, Pimentel et al. \cite{ecg_reveiw3} outlined that Right and Left ventricular dysfunction markers are independent variables that can predict COVID-19 mortality.

Ocak et al. \cite{ecg_reveiw4} performed a statistical analysis on the time duration of different ECG wave segments, ECG wave angles, different hemocyte count, hematochemical count, and patient symptoms to find out the most prominent ECG biomarker for COVID-19 prognosis and risk severity prediction. Authors pointed out that QT, QTc intervals, and frontal QRS-T angle values increase significantly in COVID-19 patients and these values increase much more as severity increases. With proper analogy and discussion, the authors proposed the `frontal QRS-T angle’ as an independent predictor of COVID-19 prognosis and severity prediction.

In a study conducted incorporating 23 French oncological institutes taking 425 RT-PCR confirmed COVID-19 positive cancer patients and 737 COVID-19 negative cancer patients aged 18 and above, Assaad et al. \cite{mortality6} discovered through a series of statistical tests that the COVID-19 positive cancer patients with either solid or hematological tumors who are getting all sorts of anticancer treatment have a 50\% 28-day mortality rate. The authors used statistical analysis (frequencies and percentages, chi-square test, median Q1–Q3, Wilcoxon test, Kaplan–Meier method, Log-rank test) on a variety of patient demographic data, tumor or cancer type, anticancer medication type with dosage and implication date, comorbidities, COVID-19 symptoms, laboratory test parameters and medications taken for other diseases. These findings reveal that the compromised immune system of cancer patients as a result of ever-evolving cancer cells, anticancer therapies and medications puts them at a greater risk of death if their respiratory, cardiac and immune systems are further harmed by the SARS-CoV-2 virus.

\subsection{Artificial Intelligence aided COVID-19 diagnostics and mortality risk prediction}
\label{2.2 Lit_rev}
The contemporary machine learning, deep learning and artificial intelligence-based studies with the most promising results are presented in this sub-section. It is important to note that, none of the studies achieved a sensitivity of 100\% which is very important to correctly classify all the positively COVID-19 infected patients, also, different datasets than this study were used.

\vspace{3mm}

\noindent \textbf{\textit{Symptoms:}} Hashmi et al. \cite{symptom_has} performed a statistical analysis on the COVID-19 early signs and symptoms collected from 10,172 laboratory-confirmed cases. The authors calculated P-value, chi-square test value, Pearson and Spearman correlation coefficients. They have observed 48.11\% sensitivity between the symptoms and COVID-19 positive cases. Finally, the authors proposed a Hashmi-Asif COVID-19 symptoms chart with corresponding probabilistic statistical values that can manually be cross-checked for COVID-19 diagnostics of a new individual. Martinez et al. \cite{symptom_mar} conducted research on COVID-19 classification based on 22 symptom features. Their proposed Voting of Decision Tree and Random Forest classifiers achieved 68.1\% accuracy, 66\% precision, 75.2\% sensitivity, 60.9\% specificity, 67\% F1-Score, and 72.8\% mean Area Under Curve (AUC).

\vspace{3mm}

\noindent \textbf{\textit{Cough Audio:}} Chetupalli et al. \cite{audio_symp1} proposed a multi-modal COVID-19 diagnostic model from patient symptoms and pre-recorded breathing, cough and speech audio data collected from the Coswara dataset \cite{audio_symp2}. Authors extracted four types of energy, thirteen types of spectral, six types of voicing, five types of percentiles, twelve types of temporal, eleven types of peaks, five types of moments, seven types of regression and two types of modulation features from the breathing, cough, speech audio data using OpenSmile \cite{audio_symp3} python toolbox followed by classification using logistic regression and support vector machine classifier. The symptom dataset had been classified using a decision tree classifier. The `multi-modal fusion score’ based model achieved an AUC of 92.40\%, sensitivity of 69\%, and specificity of 95\%.

In Project Achoo, Ponomarchuk et al. \cite{audio1} developed and launched a mobile app in iOS App Store{\textregistered} and Google Play Store{\texttrademark} that can estimate the likelihood of being COVID- 19 from patient symptoms, recorded breathing and cough data. The AI model was trained and tested on audio data collected from the publicly available datasets Coswara \cite{audio_symp2}, FSD50K \cite{audio2}, COUGHVID \cite{audio3}, Virufy \cite{audio5} and proprietary data collected from call-centers, patient recordings made by hospital staff and their mobile application. During model design, Short Time Fourier Transform and VGGish features were extracted from resampled audio data. Mel-scale Spectrogram was computed using librosa package \cite{audio6} and Cochleagram was computed using Brian2Hears package \cite{audio7} via Hann smoothed STFT. Eleven statistical features were computed for each frequency bin of the Cochleagram. Mel-Spectrogram was analyzed using two (for Mel-spectrogram and log-frequency positional encoding) channel deep convolutional neural network having custom lightweight architecture and ensemble average class probability calculated for cough-breathing audio. Cochleagram derived features and VGGish features were analyzed using the LightGBM algorithm and then the ensemble average class probability was calculated. Symptom data were analyzed using logistic regression and finally, COVID-19 positive, negative and uncertain classes were predicted from weighted output probabilities.

Early 2020 study conducted by MIT researchers developed a model using the MIT Open Voice COVID19 Cough dataset \cite{audio8} that could detect an individual’s gender and mother tongue from forced cough. They further carried out a COVID-19 screening experiment taking only English and Spanish subsets of that forced cough dataset. Mel Frequency Cepstral Coefficient calculated from the cough audio data had been sent through a deep convolutional neural network model comprising Poisson Mask layer which extracts Muscular Degradation biomarker feature and three pre-trained ResNet50’s in parallel which extracts three biomarkers namely Vocal Cord, Patient Sentiment, Lungs and Respiratory Tract features. The proposed model exhibited 98.5\% sensitivity, 94.2\% specificity on individuals diagnosed with COVID-19 via validated official test, and 100\% sensitivity, 83.2\% specificity on asymptomatic individuals. The study also revealed that the same biomarkers can differentiate between Alzheimer’s disease and COVID-19 by the power of AI models \cite{audio9}.

Mouawad et al. \cite{audio10} analyzed the recorded cough sound as well as sustained vowel `ah’ sound of a proprietary dataset for COVID-19 diagnosis as part of the Corona Voice Detection project in partnership with Voca.ai and Carnegie Mellon University. The authors calculated the Mel-Frequency Cepstral Coefficients from the time-series audio data and extracted hand-crafted Recurrence Quantification Analysis (RQA) features by using the Variable Markov Oracle (VMO) method. Their proposed XGBoost classifier machine learning model achieved an overall mean accuracy of 97\%(cough)-99\%(vowel), AUC of 84\%(cough)-86\%(vowel), a PPV measure of 78\%(cough)-69\%(vowel), and a mean F1-Score of 91\%(cough)-89\%(vowel). Nonetheless, the authors pointed out the need for the use of a multi-modal diagnostic procedure along with this sound-based approach.

\vspace{3mm}

\noindent \textbf{\textit{Blood-Test (CBC and Chemical Profile):}} Plante et al. \cite{blood5} collected blood reports of 192,779 patients admitted to 66 US hospitals just before and just after the COVID-19 pandemic outbreak. RT-PCR test results of COVID-19 positive patients had been collected on the same day as blood-test results. Their dataset included data from patients of both genders of all age groups above 20 to 80+, white-black-other skin-color races and from all the US census divisional regions including both urban and rural areas. Their XGBoost machine learning model with default hyperparameters achieved AUC score of 91\% (95\% CI 0.90-0.92), sensitivity of 95.9\%, and specificity of 41.7\%. Barbosa et al. \cite{blood4} achieved 95.16\% accuracy, 93.80\% F1-Score analyzing 24 blood features using Support Vector Machine, Random Forest, Naïve Bayes, Multilayer Perceptron machine learning classifiers. AlJame et al. \cite{blood3} achieved 99.88\% accuracy analyzing 18 blood features using ExtraTrees, Random Forest, Logistic Regression, XGBoost algorithms. 

Raihan et al. \cite{blood1} proposed a stacked ensemble machine learning model employing K-Nearest Neighbors, Support Vector Machine, XGBoost, Random Forest Classifier algorithms in the weak-learner layer and AdaBoost algorithm in the meta-learner layer of their proposed stacked ensemble model for diagnosis of COVID-19 affected patients from analysis of twenty-five hematological and hematochemical features derived from routine blood test of the patients collected from Kaggle \cite{blood_data} achieving model accuracy, precision, recall and F1-Score of 100\%. The article also elaborately described the importance of high precision and recall of such a machine learning-based infectious disease diagnostic tool exploiting the simple infectious disease model and SIR model for sustaining misclassification to a minimum through the depiction of the basic reproduction number $(R_{0})$ curve. The current research article is an extended version of the previous research work authored by Raihan et al. \cite{blood1}.

\vspace{3mm}

\noindent \textbf{\textit{Blood-Specimen Raman Spectroscopy:}} A study carried out by Yin et al. \cite{raman5} considered centrifuged blood serum and serum-saline solution of both symptomatic and asymptomatic COVID-19 positive patients, suspected patients having flue-like symptoms but COVID-19 negative and healthy individuals for Raman spectroscopic tabular-data analysis employing Support Vector Machine algorithm. The authors pre-processed the raw data through automatic-weighted least square smoothing, polynomial fitting baseline correction and total area normalization, and then the most important features were selected via ANOVA statistical test for feeding into the proposed Support Vector Machine binary classifier model. A total of 2,355 spectral data from 157 individuals were split into $70:30$ ratios for model training and test followed by external validation of the model on 300 spectral data obtained from 20 new individuals. COVID-19 positive vs suspected individual classification achieved 87\% accuracy, 89\% sensitivity and 86\% specificity. COVID-19 positive vs healthy individual classification achieved 91\% accuracy, 89\% sensitivity and 93\% specificity. Suspected vs healthy individual classification achieved 69\% accuracy, 70\% sensitivity and 66\% specificity.

Goulart et al. \cite{raman6} extracted a total of 278 (159 for COVID-19 negative and 119 for positive) Raman spectral raw data from 94 centrifuged serum samples. After that, baseline correction was performed by fitting a seventh-order polynomial to the whole spectrum range and eliminating spectral intensities attributable to fluorescence background, followed by manual cosmic ray removal and spectra normalization. Principal Component Analysis (PCA) and Partial Least Squares regression (PLS) were used to extract features, and the Kolmogorov-Smirnov normality test was used to ensure that the principal components were normalized. The study was divided into two experiments, viz. Experiment-1: COVID-19 positive vs healthy individual classification, and Experiment-2: healthy vs IgM+ vs IgG+ vs IgM+/IgG+ classification. For discovering substantial variation between the principal components, the Mann–Whitney statistical test was used in Experiment 1, and the Kruskal–Wallis statistical test was used in Experiment 2. In COVID-19 positive (IgM/IgG positive) patients, the scientists found reduced albumin but increased carotenoids, lipids, phospholipids, immunoglobulins, nucleic acids, and tryptophan. Finally, the features retrieved from both PCA and PLS were subjected to the Discriminant Analysis machine learning model. According to the findings, the PLS-DA technique outperforms the PCA-DA model. The PLS-DA method achieved 90.3\% accuracy, 84\% sensitivity and 95\% specificity using 6 principal components in the case of Experiment-1, and 88.8\% accuracy, 77.3\% sensitivity and 97.5\% specificity using 8 principal components in the case of Experiment-2.

\vspace{3mm}

\noindent \textbf{\textit{ECG Signal Image:}} Ozdemir et al. \cite{ecg_hex} investigated an ECG signal image paper-based dataset \cite{ecg_dataset} for COVID-19 and other cardiovascular disease diagnoses in 4 experimental steps wherein authors selected images only from the 0.67-25Hz bandpass filtered group of 12-lead ECG tracing paper-based images. All the image pre-processing and feature extraction were conducted using MATLAB. CNN image classification experiments were conducted using Python. During pre-processing, the authors manually segmented each channel signal containing 1 or more R-R intervals from the 12-channel ECG tracing images by `manual’ setting of rectangular frames. After that, they removed the gridlines and background in two steps (a. contrast enhancement using density mapping function via filtering input densities followed by conversion to binary images by taking `G’ channel as a reference, and b. filtering the signal from the image background using `bwareafilt’ function of MATLAB).  With these, the authors conducted their listed third experiment in which authors analyzed the pre-processed, segmented 2D ECG spectral images from 18 leads (positive polarities of 6 chest leads, both positive and negative polarities of 3 limb and 3 augmented leads) which exhibited the lowest performance among 3 other experiments. Hence for the remaining 3 experiments, they extracted features (contrast, energy, correlation, homogeneity) using Gray Level Co-Occurrence Matrix (GLCM) method which extracts signal information like periodicity and distortion, and assessed their significance through one-way ANOVA test. From segmented images, they calculated features for 18 leads (positive polarities of 6 chest leads, both positive and negative polarities of 3 limb leads and 3 augmented leads). Then again, the authors converted these 18-point features to 2D-colorful-300 DPI-dodecagonal images via Hex-axial Feature Mapping through the usage of natural two-dimensional neighbor interpolation process. These dodecagonal images were then analyzed using AlexNet, SqueezeNet, ResNet-8 and ResNet-50 CNN transfer learning models amongst which AlexNet outperformed other transfer learning models (Experiment-1). Furthermore, the authors modified the AlexNet architecture by adding one more convolutional layer having $(3\times3)$ kernel size, 256 filters, ReLu activation function in Convolutional layer and Softmax activation function in output layer followed by hyperparameter tuning using Adam optimizer that selected 0.0001 learning rate and 200 epochs (9-layer model which is their proposed model). Using this proposed model, the authors analyzed dodecagonal images (Experiment-2, 4). Overall procedures were carried out using a setup having Nvidia GeForce RTX 2080 TI GPU and 64 GB RAM using Tensor Flow 2.2 and Cuda 10.1. The analysis results (Table-\ref{table2}) are based upon `Modified stratified 5-fold cross-validation’ $(training:validation:test\ set=400:100:100)$.

\begin{table*}[htp]
\caption{Performance metrics reported by Ozdemir et al. \cite{ecg_hex}}
\centering
\begin{adjustbox}{width=\textwidth}
{\begin{tabular}{ccccccccc} \toprule

\bf{Exp.} & \bf{Data Class} & \bf{Features} & \bf{CNN model} & \bf{Accuracy} & \bf{Precision} & \bf{Recall} & \bf{F1-Score} & \bf{Specificity}
\\ \midrule

1 & D1 & F1 & C1 & 93.60\% & 91.67\% & 96.00\% & 93.76\% & 91.20\% \\
2 & D1 & F1 & C2 & 96.20\% & 94.33\% & 98.40\% & 96.30\% & 94.00\% \\
3 & D1 & F2 & C2 & 81.08\% & 79.42\% & 84.10\% & 77.81\% & 81.68\% \\
4 & D2 & F1 & C2 & 93.00\% & 90.58\% & 96.00\% & 93.20\% & 90.00\% \\

\bottomrule
\multicolumn{9}{l}{{\footnotesize{\textbf{Note:} D1 = 250 COVID-19 positive vs 250 Normal heartbeat, D2 = 250 COVID-19 positive vs 250 negative (viz. 83 normal, 83 abnormal, 84 myocardial}}} \\ 
\multicolumn{9}{l}{{\footnotesize{infarction heartbeat), F1 = Dodecagonal images from hexaxial feature mapping, F2 = 2D ECG spectral images, C1 = AlexNet, C2 = Modified AlexNet}}} \\

\end{tabular}}
\end{adjustbox}
\label{table2}
\end{table*}

Omneya Attallah \cite{ecg_biconet} proposed ECG-BiCoNet model for two-class (normal vs COVID-19 ECG traces) and three-class (normal, COVID-19, cardiac disorder ECG traces) classification on 250 images of each class from the similar ECG dataset \cite{ecg_dataset} mentioned above. At the pre-processing stage, the author resized the images as required for five transfer learning computer vision models and then performed image augmentation (flipping, translation, scaling, shearing). Then fully connected and pooling features had been extracted using five transfer learning computer vision deep-learning models namely ResNet-50, DenseNet-201, Inception-V3, Exception, and Inception-ResNet. The pooling features had been fused using the Discrete Wavelet Transform method and then integrated with the fully connected features. These features had been given the name Bi-Features. Important features were selected utilizing the Symmetrical Uncertainty method. Then, the classification had been performed using machine learning models and the best performing model was reported to be the ensemble voting classifier having Linear Discriminant Analysis, Support Vector Machine and Random Forest algorithms. The proposed ECG-BiCoNet model achieved 98.8\% accuracy-sensitivity-specificity-precision-F1Score in the case of two-class and 91.73\% accuracy-sensitivity, 95.9\% specificity, 91.9\% precision and 91.8 F1-Score in the case of three-class classification.

On the same dataset \cite{ecg_dataset}, Rahman et al. \cite{ecg_arxiv} performed two-class (Normal vs COVID19), three-class (Normal vs COVID19 vs other diseases) and five-class (all the classes given in the dataset) classification. The images had been pre-processed using gamma correction enhancement, resizing ($299\times299$ for inception, $224\times224$ for residual and dense network) followed by Z-score normalization. Augmented (translation, rotation, scaling) dataset is split into $72:8:20$ ratio as training, validation, test set and then analyzed using six deep CNN (ResNet 18/50/101, MobileNetv2, DenseNet201, InceptionV3) models with five-fold cross-validation, batch normalization, rectified linear unit, 16 mini-batch size, 15 backpropagation epochs and Adam optimizer having 0.001 learning rate, 0.2 dropout rate and 0.9 momentum update. Score-CAM heatmap was plotted to visualize the deep CNN most learning regions. For two and three class, Densenet 201 and for five-class, Inceptionv3 outperformed others with the metrics tabulated in Table-\ref{table3}.

\begin{table*}[htp]
\caption{Performance metrics reported by Rahman et al. \cite{ecg_arxiv}}
\centering
\begin{adjustbox}{width=\textwidth}
{\begin{tabular}{cccccccc} \toprule

\bf{Data Class} & \bf{Algorithm} & \bf{Confidence Intervals} & \bf{Accuracy} & \bf{Precision} & \bf{Sensitivity} & \bf{F1-Score} & \bf{Specificity}
\\ \midrule

2-Class & Densenet201 &  & $99.10\pm0.44\%$ & $99.11\pm0.43\%$ & $99.10\pm0.44\%$ & $99.09\pm0.44\%$ & $96.90\pm0.80\%$ \\
3-Class & Densenet201 & $95\%$ & $97.36\pm0.74\%$ & $97.40\pm0.74\%$ & $97.36\pm0.74\%$ & $97.36\pm0.74\%$ & $97.93\pm0.66\%$ \\
5-Class & InceptionV3 &  & $97.83\pm0.67\%$ & $97.82\pm0.67\%$ & $97.83\pm0.67\%$ & $97.82\pm0.67\%$ & $98.86\pm0.49\%$ \\

\bottomrule

\end{tabular}}
\end{adjustbox}
\label{table3}
\end{table*}

Despite image augmentation has some benefits to a certain extent but after that limit and in the case of ECG image classification, it has pitfalls in classification using Deep CNN since the augmentation method synthetically duplicates the same image data increasing that particular class weight which mimics the unrealistic situation. Anwar et al. \cite{ecg_aug} investigated the same dataset \cite{ecg_dataset} resizing images into $300\times300$ pixels followed by four augmentations (flipping, crop, perspective distortion, contrast and gamma change). Then image classification was conducted using EfficientNet B3 deep learning model with a learning rate 0.0001, Adam and AdamW optimizer, 5-fold cross-validation focusing on the effect of augmentation. The authors concluded with classification obtaining 81.8\% accuracy, 77.6\% F1-Score without image augmentation and the performance deteriorated to 76.4\% accuracy, 76.8\% F1-Score with augmentation. According to the authors, core information about the diseases lies in the PQRST pattern of the ECG signal. If the images are flipped, cropped, tilted, rotated, elevated/lowered for augmentation, it will cause PQRST pattern distortion falsifying the pathological phenomena which adversely affect model performance.

\vspace{3mm}

\noindent \textbf{\textit{Mortality Risk Prediction:}} Pourhomayoun et al. \cite{mortality1} selected 57 important features for mortality risk prediction out of a total of 112 features obtained from 307,382 labeled samples among 2,670,000 lab-confirmed COVID-19 patients from 146 countries. The authors applied a couple of different filter and wrapper feature selection techniques for important feature selection from 3 categories of features, viz. patient symptoms, comorbidities and patient demographics. Their best performing algorithm Artificial Neural Network achieved 89.98\% accuracy. K-Nearest Neighbor and Support Vector Machine also achieved 89.83\% and 89.02\% accuracy respectively. 

Hoon Ko et al. \cite{mortality2} proposed EDRnet (Ensemble learning model based on Deep Neural Network and Random Forest) model for COVID-19 infected patient mortality rate prediction from 28 blood features along with gender and age. The model was trained on 361 samples. Their proposed EDRnet model tested on 106 patient data achieved an accuracy of 93\%, sensitivity of 100\% and specificity of 91\%.

Mahdavi et al. \cite{mortality3} used 37 features including both invasive and non-invasive features for mortality risk prediction from 492 patients (66.1\% male, 33.9\% female) admitted to the hospital cohort. Non-invasive features include data from the sub-categories of patient demographic, symptoms, comorbidities, blood pressure (systolic and diastolic), pulse rate, respiratory rate, oxygen saturation (SpO$_{2}$) and body temperature. Micro-invasive features include data from the sub-categories of complete blood count, blood coagulation, biochemistry and blood gas test. The authors analyzed the micro-invasive, non-invasive and joint features using three Support Vector Machine models and provided a comprehensive comparison of these three models based on feature types. Experiments revealed that the joint features outperform others securing 92\% AUC, 81\% sensitivity and 91\% specificity. However, the non-invasive features secured the second-best position in COVID-19 patient mortality risk prediction according to Mahdavi et al. 

Subudhi et al \cite{mortality5} developed a machine learning model after scrutinizing 18 different algorithms for predicting both ICU admission possibility within 5 days after presenting to the Emergency Department (ED) of the hospital cohort due to COVID-19 positive confirmation and mortality risk of the patient within 28 days followed by the ICU admission date. The authors considered a total of 56 features on patient demographic data, medication usage, history of past illness, clinical examination data, blood count and hematochemical profile for the machine learning model input. ICU admission possibility prediction training dataset comprised of data obtained from 486 ICU-admitted patients and 486 non-ICU admitted patients. Mortality risk prediction training dataset included data obtained from 344 corpses (dead patients in ICU) before death and 344 ICU-admitted patients who were still alive after 28 days of ICU admission. In the ICU admission experiment, the best-performing machine learning models reported are Bagging Classifier, Gradient Boosting Classifier and Random Forest Classifier all achieving an F1-Score of 81\% with 95\% Confidence Interval. In case of the mortality risk prediction experiment, Linear Discriminant Analysis algorithm performed the best securing 88\% F1-Score having 95\% Confidence Interval. The authors also calculated and tabulated the SHAP values along with other statistics of the most important features to identify risk factors and critical variables for predicting ICU admission and mortality risk. A positive SHAP score represents that the corresponding variable raises the risk of ICU admission and mortality. The authors outlined that C-reactive Protein, Lactate Dehydrogenase and Oxygen Saturation are the most crucial features for ICU admission models, whereas the Estimated Glomerular Filtration Rate less than 60 milliliters per minute per 1.73 square meters, Neutrophil and Lymphocyte percentages are the most crucial features for predicting mortality risk. 

Nemati et al. \cite{mortality4} used the patient demographic data from 1,182 COVID-19 patients to assess the accuracy of multiple survival analysis methodologies along with the prediction of discharge time of hospitalized patients by harnessing both statistical and machine learning algorithms. Gradient Boosting (GB), Cox Proportional Hazard (CoxPH), Coxnet, and radial basis kernel function SVM achieved an accuracy of around 71\%. GB was the most accurate, while IPCRidge performed the worst with an accuracy of 49\%. Irfan et al. \cite{mortality7} carried out a survey on both machine learning and deep learning models for COVID-19 mortality risk prediction and observed that deep learning models can predict mortality risk more accurately even with fewer features.

\section{Methodology}
\label{Methodology}

\begin{figure}[ht]
\centering 
\resizebox*{\textwidth}{!}{\includegraphics{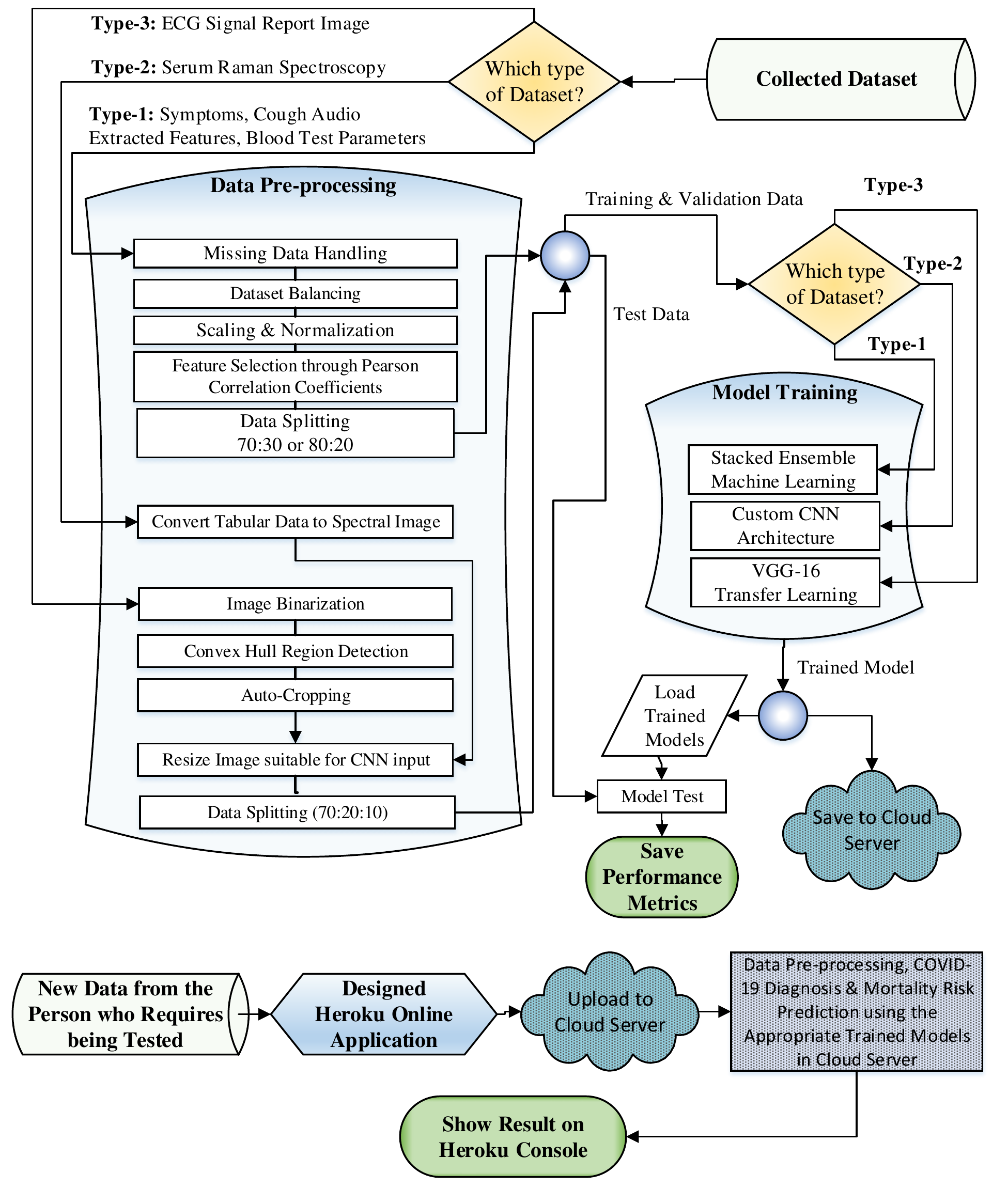}}
\caption{{Overall methodologies represented as block diagram.}}
\label{fig2}
\end{figure}

Rapid point-of-care diagnosis of COVID-19 infected patients, subsequent mortality risk prediction of the patients, and isolating them from healthy individuals are of paramount importance. This article addresses both COVID-19 diagnosis and mortality risk prediction using machine learning and deep learning approaches along with a prototype multimodal online app implementation. This study proposes a multimodal solution for COVID-19 diagnosis using machine learning and deep learning techniques based on suspected individual’s symptoms, cough sound, hematological data, Raman spectroscopy image of blood serum and ECG signal image. Only hematological data is utilized for the mortality risk prediction of COVID-19 positive patients. Symptoms and cough audio are readily available to the individuals ready for the test. The blood profile along with the Raman spectroscopy image and ECG signal image can be obtained from any nearby pathological centers. It is assumed that a patient has already tested positive for COVID-19 in order to calculate the mortality risk. The COVID-19 diagnostic test using the multimodal online application can be done by the individual thyself while staying at home or at any healthcare/diagnostic center nearby even having a low resource setting. This section outlines the design and development process of this study in chronological sequence. The overall procedure is graphically represented as block diagram in Figure-\ref{fig2}.

\subsection{Dataset Overview}
\label{Dataset Overview}
Eight distinct datasets regarding Symptoms \cite{symp_data}, Cough Audio \cite{cough_data_1, cough_data_2}, Routine Blood Test \cite{blood_data}, Raman Spectroscopy Data \cite{raman5}, ECG Signal Image \cite{ecg_dataset}, Mortality Risk Prediction from Hematological Data \cite{ mortality2, mortality3} have been considered during the development of this research work. 

A total of 21,235 samples, each with 5 features have been considered for analysis in this study from the symptoms dataset \cite{symp_data}. Recorded Cough Audio samples are collected from two different datasets \cite{cough_data_1, cough_data_2} which are then merged together. Finally, a total of 586 audio samples are selected for 7 hand-crafted feature extraction and further analysis. For the COVID-19 diagnosis from the hematological data \cite{blood_data} experiment, 603 samples with 25 blood-related attributes are considered. 

Both centrifuged blood serum and serum-saline solution had been used by Yin et al. \cite{raman5} for obtaining the Raman Spectroscopic data. The tabular dataset provided Raman spectral data of 309 samples having 900 columns corresponding to the wavelength of light in units of Raman Shift per Centimeter. This study plotted each row of the tabular dataset for converting them into a dataset of 309 raw images.

\clearpage

To the best of the authors' knowledge, ECG signal image dataset \cite{ecg_dataset} is the first ever and only available publicly shared paper-based ECG image dataset which includes 12 lead ECG tracing of individuals involving COVID-19 infected (250), Myocardial Infarction (77), Previous History of Myocardial Infarction (203), Normal Heartbeat (859), Abnormal Heartbeat (548). The data is accumulated using EDAN SERIES - 3 devices of 500 Hz sampling rate from 3 different institutions of Pakistan and the paper-based ECG image data selection process undergone through several months of manual review by medical professors supervised by senior medical professionals. Since the data had been collected from 3 different institutions, the report images are diverse. All the images of different disease classes are not of the same resolution and the report pages are not of a similar standard. Some non-COVID ECG tracing reports included the names of possibly diagnosed heart diseases enlisted at the top of the report and few reports have some other text markers at the top of the report page. All the COVID-19 positive ECG tracings are placed at the top of portrait paper but all other ECG reports were prepared on landscape paper. Since the area inside the boundary of tracing regions is the region of interest for all images and all tracings had been collected using a similar type of ECG machine from the same manufacturer, the images can be analyzed after some image pre-processing (cropping and gridline-removal) to make all the reports of similar standard and resolution excluded of any unnecessary regions and report texts. However, the number of ECG signal-tracing rows in a report, the signal-tracing order of the different ECG-leads on the printed report (especially in the case of PMI target class), and the ECG device bandpass frequency are seen as dissimilar (Supplementary Material - \ref{sup_mat} (b)) in the case of many reports present as dispersed chunks within the bulk of the images of the entire huge dataset that could not be manually excluded or corrected and believed to cause a lot of redundancy and shortcut-learning (similar to  DeGrave et al. \cite{xray3}) if employed for 5-class classification. For these reasons, this study confined the experiment in classifying COVID-19 vs Normal Heartbeat using selected 1,109 ECG signal images from the entire dataset. The images are then converted into binary and auto-cropped to remove the white space and text labels by implementing a convex hull.

The Mortality Risk Prediction of COVID-19 positive patients has been analyzed in two experiments. The first one is using 7 Routine Blood Test parameters of 361 patients from the dataset published by Hoon et al. \cite{ mortality2}, and the other one is using 9 Routine Blood Test parameters of 628 patients from the dataset published by Mahdavi et al. \cite{mortality3}.

Information regarding the dataset samples, features, pre-processing and applied algorithms for all the experiments conducted in this study are tabulated in Table-\ref{table4}.


\begin{sidewaystable*}[htp]
\caption{Overview of the datasets and applied algorithms of all the experiments conducted in this study.}
\centering
\begin{adjustbox}{width=\textwidth}
{\begin{tabular}{ccclllllllcll} \toprule

{\bf{Exp.}} & & {\bf{Dataset and Purpose}} & &  {\bf{Feature and Samples}}  & & {\bf{Features
}} & & {\bf{Data Preprocessing}} & & {\bf{Alg$^{thm}$ Type}} & & {\bf{Applied Algorithms}}
\\ \cmidrule{1-1} \cmidrule{3-3} \cmidrule{5-5} \cmidrule{7-7} \cmidrule{9-9} \cmidrule{11-11} \cmidrule{13-13}

\multirow{4}{*}{1} && Symptoms \cite{symp_data}  && \#Samples: 21,235   && Headache, Fever, && MVH: KNN Imputer && Stacked && BL-1: RFC \\
                   && (COVID-19                  && \#Features: 5 && Cough, Sore throat and && DB: SMOTE &&    Ensemble && BL-2: XGBoost \\
                   && Diagnosis)                 && FET: Dataset Provided  && Shortness of breath && FSN: Yes && Machine && BL-3: SVM \\
                   &&                            && FDT: Tabular  && && DS: $70:30$ && Learning &&  MetL: NB \\\\
                   
\multirow{4}{*}{2} && Recorded Cough  && \#Samples: 586   && Minimum, Maximum, && MVH: KNN Imputer && Stacked && BL-1: RFC \\
                   && Audio \cite{cough_data_1, cough_data_2} && \#Features: 7 && Mean, Standard deviation, && DB: SMOTE &&    Ensemble && BL-2: XGBoost \\
                   && (COVID-19                  && FET: Hand-Crafted  && Skewness, Kurtosis and && FSN: Yes && Machine && BL-3: DT \\
                   && Diagnosis)                           && FDT: Tabular  && Dominant Frequency && DS: $70:30$ && Learning &&  MetL: LR \\\\
                   
\multirow{8}{*}{3.1} &&   &&    && \textbf{Demography:} Age &&  && && \\
                   &&  &&  && \textbf{CBC:} Hemoglobin, RBC, HCT, MCV, MCH, MCHC, &&  && && \\
                   && Blood-Test \cite{blood_data} && \#Samples: 603  && RDW, TWBC, Neutrophils, Eosinophils, Basophils, && MVH: KNN Imputer && Stacked && BL-1: RFC \\
                   && (COVID-19 && \#Features: 25  && Lymphocytes, Monocytes, Platelets, MPV && DB: SMOTE && Ensemble && BL-2: XGBoost \\
                   && Diagnosis) && FET: Dataset Provided && \textbf{Chemical Profile:} Albumin, Sodium, Potassium, && FSN: Yes && Machine && BL-3: SVM \\
                   &&  && FDT: Tabular && Alanine transaminase, Aspartate transaminase, Hs-CRP, && DS: $70:30$ && Learning &&  MetL: NB \\
                   && && && Creatinine and Urea    && && && \\
                   && && && \textbf{Coagulation:} PT  && && && \\\\
                   
\multirow{4}{*}{3.2} && Blood-Test \cite{blood_data}  && \#Samples: 603   &&  && MVH: KNN Imputer && Stacked && BL-1: RFC \\
                   && (COVID-19 && \#Features: 5 && \textbf{Demography:} Age && DB: SMOTE &&    Ensemble && BL-2: XGBoost \\
                   && Diagnosis)                  && FET: Dataset Provided  && \textbf{CBC:} TWBC, Eosinophils, Monocytes, Platelets  && FSN: Yes && Machine && BL-3: KNN \\
                   &&                            && FDT: Tabular  &&  && DS: $70:30$ && Learning &&  MetL: NB \\\\
                   
\multirow{4}{*}{4} && Blood-Serum Raman  && \#Samples: 309   &&  &&  && Deep &&  \\
                   && Spectroscopy \cite{raman5} && Tabular to Image && Deep Learning Inherent Automatic Feature && DS: $70:20:10$ &&    Learning && Custom CNN \\
                   && (COVID-19 && Conversion.  && Extraction &&  && Computer && Architecture \\
                   && Diagnosis) && FDT: Signal Image  &&  &&  && Vision &&   \\\\
                   
\multirow{4}{*}{5} && ECG Signal \cite{ecg_dataset}  && \#Samples: 1,109   && Deep Learning Inherent Automatic Feature && Binarization, Auto-cropping && Deep &&  \\
                   && (COVID-19 && FDT: Signal Image && Extraction && to remove the white space and &&    Learning && VGG16 \\
                   && Diagnosis) &&   &&  && text labels by implementing && Computer &&  \\
                   &&  &&   &&  && Convex Hull. DS: $70:20:10$ && Vision &&   \\\\
                   
\multirow{4}{*}{6.1} && Blood-Test \cite{mortality2}  && \#Samples: 361   && \textbf{CBC:} Neutrophils, Lymphocytes, Monocytes, Platelets && MVH: KNN Imputer && Stacked && BL-1: RFC \\
                   && (COVID-19 && \#Features: 7 && \textbf{Chemical Profile:} Albumin, Hs-CRP && DB: SMOTE &&    Ensemble && BL-2: XGBoost \\
                   && Mortality Risk                  && FET: Dataset Provided  && \textbf{Coagulation:} PT && FSN: Yes && Machine && BL-3: SVM \\
                   && Prediction) && FDT: Tabular  &&  && DS: $80:20$ && Learning &&  MetL: RFC \\\\
                   
\multirow{5}{*}{6.2} && Blood-Test \cite{mortality3}  && \#Samples: 628   && \textbf{Demography:} Age && MVH: KNN Imputer && Stacked && BL-1: RFC \\
                   && (COVID-19 && \#Features: 9 && \textbf{CBC:} MCHC, RDW, TWBC && DB: SMOTE &&    Ensemble && BL-2: XGBoost \\
                   && Mortality Risk                  && FET: Dataset Provided  && \textbf{Chemical Profile:} BE && FSN: Yes && Machine && BL-3: SVM \\
                   && Prediction) && FDT: Tabular  && \textbf{Coagulation:} PT, PTT && DS: $80:20$ && Learning &&  MetL: RFC \\
                   && && && \textbf{Others:} RR, SpO$_{2}$ && && && \\\\

\bottomrule

\multicolumn{13}{l}{{\small{\textbf{\underline{Accronyms:}}} \footnotesize{\underline{FET:} Feature Extraction Type, \underline{FDT:} Final Data Type, \underline{MVH:} Missing Value Handling, \underline{DB:} Dataset Balancing, \underline{FSN:} Feature Scaling and Normalization, \underline{DS:} Dataset Splitting, \underline{BL:} Base Learner, \underline{MetL:} Meta Learner, }}}\\

\multicolumn{13}{l}{{\footnotesize{\underline{RFC:} Random Forest Classifier, \underline{SVM:} Support Vector Machine, \underline{NB:} Na{\"{\i}}ve Bayes, \underline{DT:} Decision Tree, \underline{LR:} Logistic Regression, \underline{KNN:} K-Nearest Neighbor, \underline{CNN:} Convolutional Neural Network}}}\\

\multicolumn{13}{l}{{\small{\textbf{\underline{Hematological Accronyms:}}} \footnotesize{\underline{CBC:} Complete Blood Count, \underline{RBC:} Red Blood Cells, \underline{HCT:} Hematocrit, \underline{MCV:} Mean Corpuscular Volume, \underline{MCH:} Mean Corpuscular Hemoglobin, \underline{MCHC:} Mean Corpuscular Hemoglobin Concentration, \underline{BE:} Base Excess,}}}\\

\multicolumn{13}{l}{{\footnotesize{\underline{RDW:} Red Cell Distribution Width, \underline{TWBC:} Total White Blood Cells, \underline{MPV:} Mean Platelet Volume, \underline{Hs-CRP:} Hypersensitive c-reactive protein, \underline{PT:} Prothrombin Time, \underline{PTT:} Partial Thromboplastin Time, \underline{RR:} Respiratory Rate, \underline{SpO$_{2}$:} Oxygen Saturation}}}\\

\end{tabular}}
\end{adjustbox}
\label{table4}
\end{sidewaystable*}

\subsection{Dataset Pre-processing}
\label{preprocessing}
At first, categorical features present in the datasets, e.g. textual data, were converted into numeric values using the One Hot Encoding technique.

\vspace{3mm}

\noindent \textbf{\textit{Missing Data Handling:}} The capability of statistical estimation of a machine learning model is hampered due to missing values in a dataset that may occur even in a controlled and well-scrutinized study. Until this time, several data imputation techniques have been invented among which the K-Nearest Neighbors (KNN) imputer \cite{imputer} has been used in this study for the machine learning models with $k=5$ number of neighbors to handle null data in the corresponding dataset. The KNN imputer imputes the missing values with the mean value of k-nearest neighbors to the missing valued training sample. The nearest distance is identified via calculation of the Euclidean Distance metric between the other available feature space of the missing valued sample and the rest of the samples of the dataset.

\vspace{3mm}

\noindent \textbf{\textit{SMOTE Analysis:}} Imbalanced distribution of target class in the dataset poses a challenge in developing an effective model irrespective of various classifiers since the predictive model tends to bias more toward the target class of high frequency. To ameliorate this raised concern, one of the most popular data oversampling algorithms known as `Synthetic Minority Over-sampling Technique (SMOTE)' \cite{smote} has been employed in the respective machine learning analyses of this study. SMOTE technique performs synthetic data point fabrication utilizing the working principle of k-nearest neighbors. 

\vspace{3mm}

\noindent \textbf{\textit{Data Splitting:}} All of the COVID-19 diagnostic machine learning related tabular datasets have been split maintaining $70:30$ and mortality risk prediction datasets have been split maintaining an $80:20$ $\ training:test$ ratio. The deep learning computer vision related datasets have been split maintaining a $70:20:10$    $\ training:validation:test$ ratio. 

\vspace{3mm}

\noindent \textbf{\textit{Feature Scaling:}} Extracted features from cough audio signal, features from symptoms and blood sample analysis lies in a range of different numerical values. To work with these data stemmed from real-world problems, it's imperative to bring all the data in a certain range in order to save computation power and make the calculation faster. Therefore, the standard scalar (\textit{sklearn.preprocessing.StandardScaler}) \cite{scaler} method is used in case of all machine learning related datasets to standardize the features. 

\vspace{3mm}

\begin{figure}[ht]
\centering 
\resizebox*{\textwidth}{!}{\includegraphics{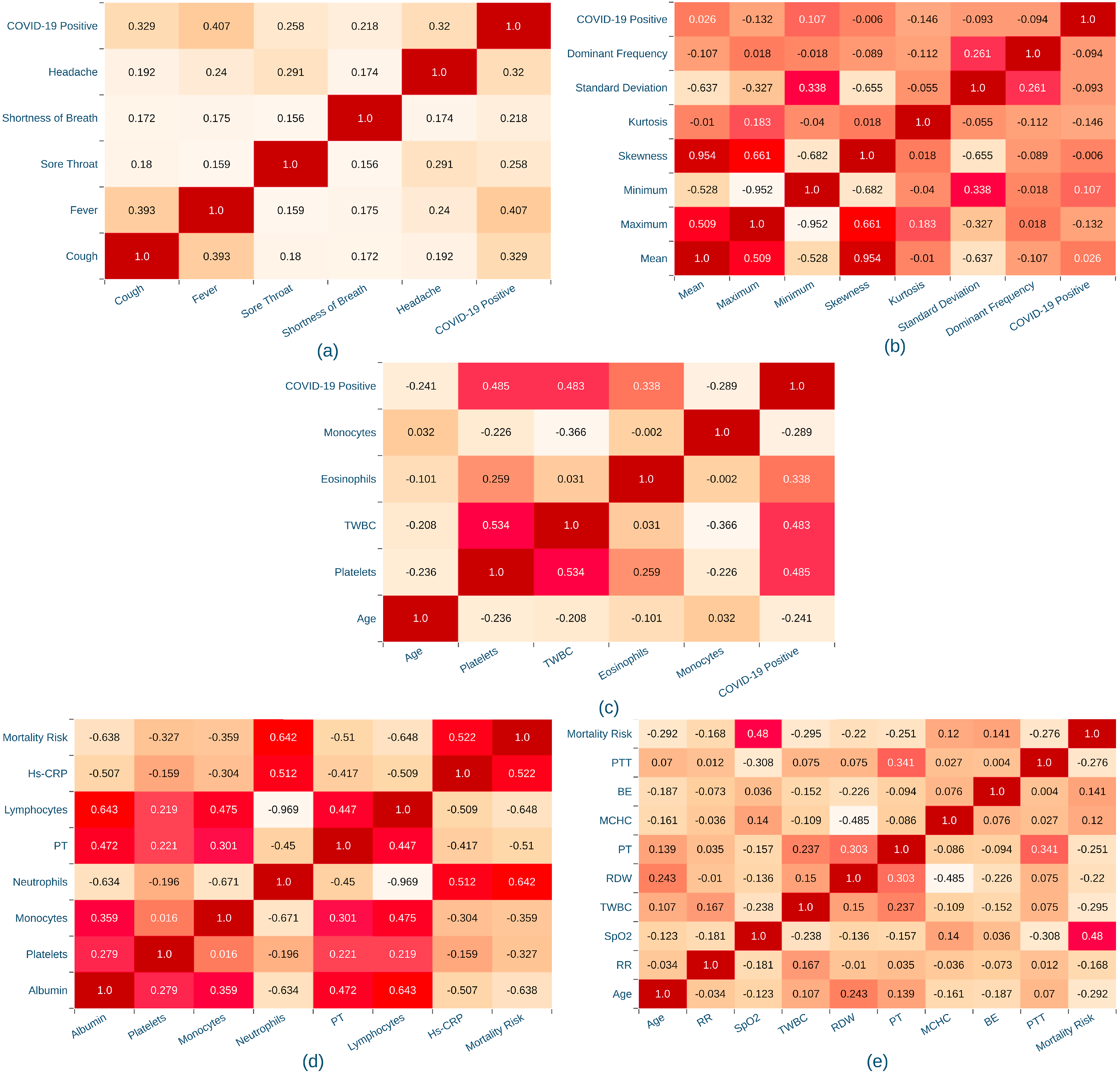}}
\caption{{Pearson Correlation Coefficients of Dataset Features: (a) Symptoms (b) Cough Audio (c) Blood Test (5 features) for COVID-19 diagnosis (d) Blood Test (7 features) for Mortality Risk prediction (e) Blood Test (9 features) for Mortality Risk prediction.}}
\label{fig3}
\end{figure}

\noindent \textbf{\textit{Feature Correlation:}} Pearson Correlation Coefficients \cite{pearson} of the machine learning related datasets have been computed and plotted (Figure-\ref{fig3}) in order to explore the relationship between attributes and target variable, multi-collinearity analysis and feature selection. Since blood tests for too many parameters will cost excessive test fees to the patients and will also be a burdening issue for pathological centers, this study aimed to find out the most important and highly correlated features with respect to the target outcome. Hence, only 25 hematological features have been initially selected for analysis in this study from the original dataset \cite{blood_data} containing a total of 110 clinical features. Then, most important 5 features (out of 25) have also been analyzed upon scrutinizing the correlation coefficients from those 25 features for COVID-19 diagnosis. Similarly, through scrutinization of the correlation coefficients, 7 most important features have been selected out of 28 features from the dataset published by Hoon Ko et al. \cite{mortality2} and 9 most important features have been selected out of 37 features from the dataset published by Mahdavi et al. \cite{mortality3} in case of Mortality Risk prediction related datasets.

\clearpage

\begin{figure}[ht]
\centering 
\resizebox*{\textwidth}{!}{\includegraphics{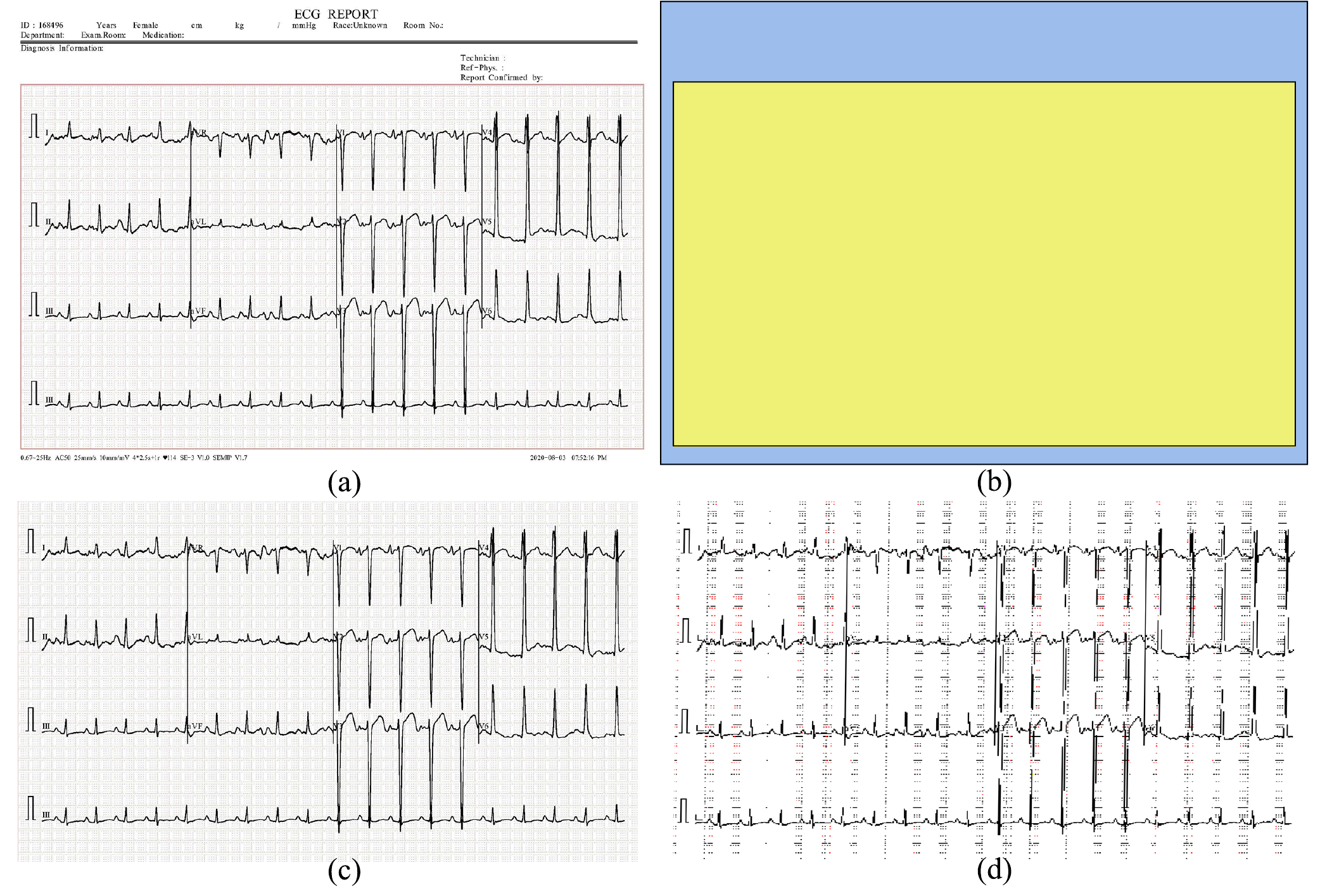}}
\caption{{ECG signal image pre-processing (a) Original Image (b) Convex Hull detection in the binary image (c) Auto-cropping around the Convex Hull (d) Image after filtering while inside the deep-learning computer vision model.}}
\label{fig4}
\end{figure}

\noindent \textbf{\textit{ECG Image Data Pre-processing:}} To get rid of the anomalies in the ECG signal image reports described in section-\ref{Dataset Overview}, this study converted all the images into binary (black and white mode). Then, the Convex Hull region is programmatically detected which encloses only the signal tracings region of the report image. Finally, that region is auto-cropped to remove the white space and text labels. All the images are rescaled into $224\times224$ pixels to make them of similar resolution and suitable for deep-learning model input. It is observed that when the input images are passed through several Convolution and Max-Pooling operations with a certain filter-size at each operation, the computer vision deep learning model can extract the ECG signal portion as features such that the grid lines are almost removed and the signal tracings are retained as binary image format. The process is depicted in Figure-\ref{fig4}.

\subsection{Machine Learning Algorithms}
\label{ml}
The following is a summary of the machine learning algorithms applied to the various tabular datasets of this study. 

\vspace{3mm}

\noindent \textbf{\textit{Logistic Regression (LR):}} The logistics regression algorithm is a classification algorithm that is built upon the concept of probability. In this algorithm, the observations are assigned to a discrete set of classes. To map the values of probabilities, a logistic function called the sigmoid function takes a set of attributes as input and the function maps any real value into another value between 0 and 1 for classifying the input as a positive or negative class based on the position of output against the threshold line \cite{ml_algorithm}. Default hyperparameters have been used for LR with random\_state = 0 in this study. 

\vspace{3mm}

\noindent \textbf{\textit{K-Nearest Neighbors (KNN):}} KNN is a lazy learning algorithm that computes the statistical distance, e.g. Euclidean distance, of a test datapoint from each of the training datapoint. The algorithm predicts the class of the test data based on the target class of k-nearest distant training datapoints \cite{ml_algorithm}. Following parameters were selected in this research work for KNN: n\_neighbors = 5, metric = `minkowski', p = 2. 

\vspace{3mm}

\noindent \textbf{\textit{Support Vector Machine (SVM):}} SVM classifier groups the training datapoints in n-dimensional space according to their target class and finds the critical edge points called support vectors. Then it draws hyperplanes touching the support vectors and gets the line separating the hyperplanes at maximum distance. The algorithm predicts the target class of a test datapoint based on the area that falls in \cite{ml_algorithm}. `rbf' kernel with random\_state = 0 has been used for SVM in the analyses of this study. 

\vspace{3mm}

\noindent \textbf{\textit{Na{\"{\i}}ve Bayes (NB):}} Using the Bayes theorem, this classifier measures the probability of each target class assuming all the features are not correlated with each other and finally predicts the class with the highest probability \cite{ml_algorithm}. GaussianNB(.) with default parameters have been used for this study. 

\vspace{3mm}

\noindent \textbf{\textit{Decision Tree (DT):}} It is a probabilistic model wherein the data sample is repeatedly split based on certain parameters. Starting from the root node it calculates the information gain and entropy of each feature and selects the feature corresponding largest information gain or smallest entropy. The decisions are split based on a decision node which acts as a condition for each node. That's how the algorithm forms a `tree-structured flow chart' and finally predicts the target class via a top-down attribute search approach. The leaves of the tree define the final outcomes of the algorithm \cite{ml_algorithm}. Following parameters have been selected for DT: criterion = `entropy', max\_leaf\_nodes = 300, random\_state = 0. 

\vspace{3mm}

\noindent \textbf{\textit{Random Forest Classifier (RFC):}} This classifier combines a multitude of Decision Trees using the `bagging' method for eliminating the pitfalls (data overfitting and low variance) associated with DT. All of the individual decision trees in a random forest are trained using different random subsets of the original training dataset. A typical implementation of the RFC algorithm uses majority voting as its aggregating method for the final result \cite{ml_algorithm}. Following parameters have been selected for RF in this research work: n\_estimators = 1500, criterion = `gini', random\_state = 0. 

\vspace{3mm}

\noindent \textbf{\textit{Extreme Gradient Boosting (XGBoost):}} This algorithm performs distributed and parallel computing using an ensemble of various classifiers using distributed and optimized gradient boosting techniques. Parallel computing makes the process more efficient and faster \cite{ml_algorithm}. Following parameters have been selected for XGB in this research work: n\_estimators = 25, max\_depth = 15, subsample = 0.7. 

\vspace{3mm}

\noindent \textbf{\textit{Stacked Ensemble Machine Learning:}} While applying different machine-learning or deep-learning algorithms with the best hyperparameters after tuning them to the dataset, it might have been observed that a couple of algorithms are achieving nearly the highest test benchmark. Combining the predictions of all those nearly best algorithms along with the best performing algorithm often enhances the performance even more. If those probabilistic predictions are fed as input to another layer of combined algorithms to produce the final classification output, then the second layer is said to be stacked atop the first layer. Similarly, another layer can also be stacked atop the second layer and so on. This task of multiple-algorithm-combination and stacking them atop is achieved through the process called Stacked Generalization and the final model is called the Stacked Ensemble Machine Learning model. The base layer is termed as the Weak-Learner or Base-Learner layer and the $n_{th}$ successive layer atop is termed as the Level-n Meta-Learner layer. In the proposed stacked ensemble model, this study used a double-layer stacked ensemble model (Figure-\ref{fig5}(a) and Table-\ref{table4}). Weak-learner makes preliminary predictions having relatively less performance but merging a few weak learners' predictions as a training dataset for meta-learner improves the overall system. Many researchers nowadays are using this technique to push their model performance to its epitome. \cite{stacked_ml}

\subsection{Deep Learning Computer Vision Algorithms}
\label{dl}
\begin{figure}[ht]
\centering 
\resizebox*{11cm}{!}{\includegraphics{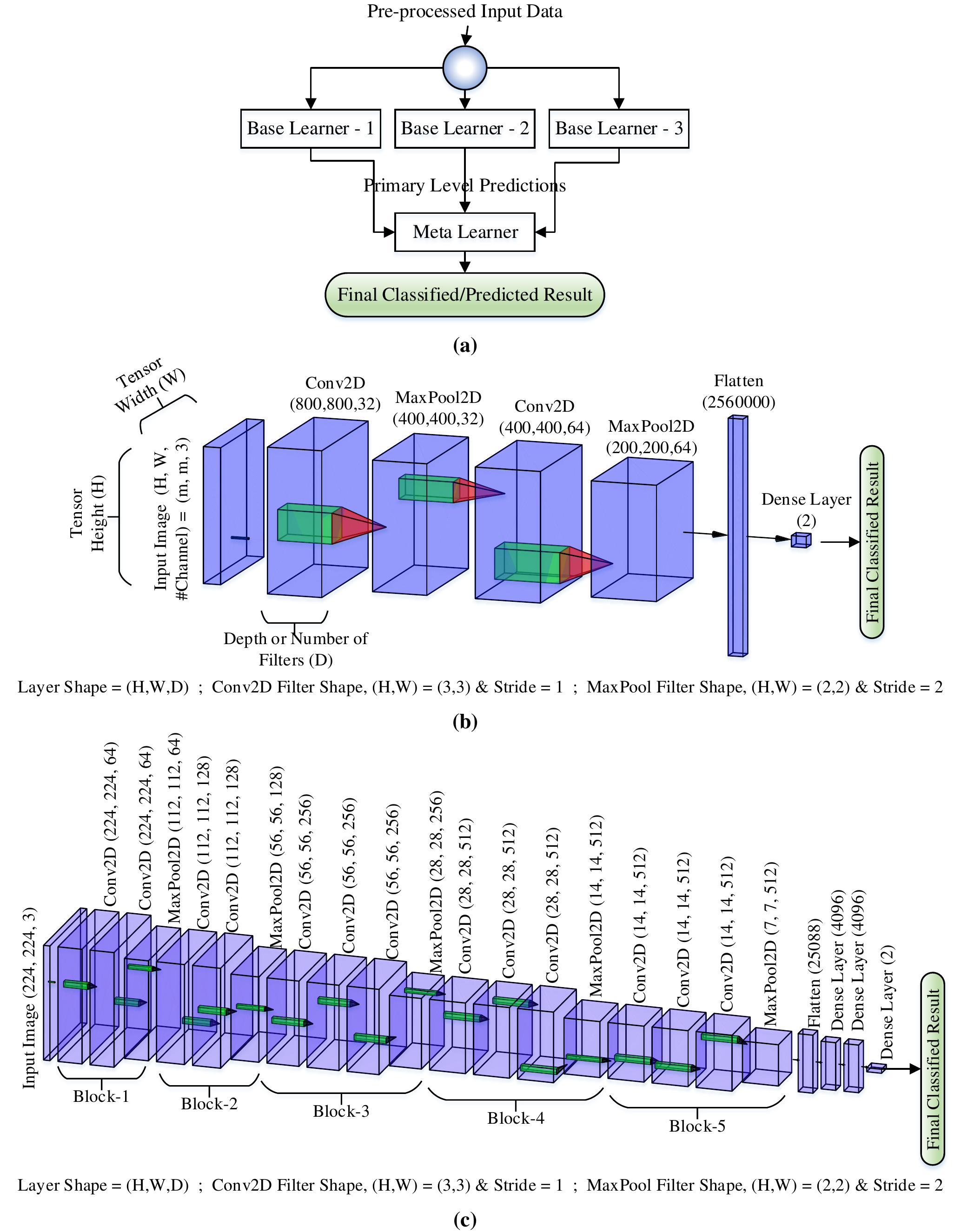}}
\caption{{(a) Stacked Ensemble Machine Learning Architecture  (b) Custom CNN Architecture  (c) VGG-16 Transfer Learning Architecture}}
\label{fig5}
\end{figure}

The Raman Spectral tabular data obtained from the centrifuged blood serum Raman spectroscopy analysis have been converted to images and then classified using a custom Convolutional Neural Network (CNN) architecture (Figure-\ref{fig5}(b)). A Convolutional Neural Network (CNN) \cite{CNN_review1, CNN_review2} is a type of human-brain interconnected neuron architecture inspired deep learning technique that uses the convolution mechanism to anatomize visual imagery and distinguish patterns.

For the ECG signal image dataset, this study used a variant of transfer learning CNN architecture called the Visual Geometry Group-16 (VGG-16) model (Figure-\ref{fig5}(c)) with the final modified to output 2 classes (COVID-19 positive or normal heartbeat). The VGG-16 \cite{CNN_review1} is popular for its uniformity in Convolutional Kernel size of $3\times3$ and Maxpool Kernel size of $2\times2$. All the Convolutional operations use a stride of 1 and all the Maxpooling operations use a stride of 2 while the padding keeps the image size same across convolutional layers.

Both the models have been trained for 25 epochs using the Adam optimizer with a 0.0001 learning rate and Categorical Cross-entropy as the loss-metric. The CNN architecture and its transfer-learning variations operate on the basis of three layers, which are outlined below. 

\vspace{3mm}

\noindent \textbf{\textit{Convolutional Layer:}} This is the premier component of the CNN as the majority of the computations are performed here. The layer aims at extracting the high-level to gradually mid-level and low-level features such as edges, textures, color, etc. from the input image. The convolution process detects features using a kernel or a filter that has the size of a $3\times3$ matrix in case of both the proposed custom architecture and VGG-16 Conv2D layer. Calculation of a dot product between the input pixels and the filter is done upon applying the filter to the image part. This dot product is then fed into an output array. The kernel sweeps the full image by repeatedly shifting through strides. The final output from the series of dot products from the input and the filter is known as a feature map or a convolved feature. The input image matrix \textbf{I} is convolved with the 2D kernel \textbf{K} of $m \times n$ size. The convolution $\textbf{I*K}$ produces the output feature map F as follows. 
\begin{align}
F(i,j)=(I*K)(i,j)=\sum\limits_{m}{\sum\limits_{n}{I(i+m)(j+n)K(m,n)}}
\end{align} 

The piecewise activation function called Rectified Linear Unit (ReLU) is applied to the feature map of each convolutional layer. The function outputs the input directly if it is positive, otherwise, outputs zero. The mathematical representation of this function is $ f(a) = max(0,a)$ .

\vspace{3mm}

\noindent \textbf{\textit{Pooling Layer:}} Down-sampling along the spatial dimension of the convolved feature is done in this layer. This dimensionality reduction leads to the requirement of less computation power in the processing of data. Out of different pooling methods, this study used the Max-Pooling layer. Max pooling collects the maximum value from the region covered by the kernel. The pooling operation sweeps the kernel across the entire image by repeatedly shifting through strides. The layer thus succeeds in extracting rotational and positional invariant dominant features. Pooling is crucial in CNN as it extracts features, controls overfitting, and boosts efficiency. 

\vspace{3mm}

\noindent \textbf{\textit{Fully Connected Layer or Dense Layer:}} In the fully-connected layer, each node in the output layer is connected to every node in the preceding layer through a direct connection. Outputs obtained through the previous layers and their different filters are used for the classification task. The class score is calculated by forwarding the output of the final fully connected layer to an activation function named Softmax. The function evokes a probability ranging between 0 and 1 using the following equation.
\begin{align}
{{Z}_{i}}=\frac{{exp\left({z}_{i}\right)}}{\sum\limits_{k}{{exp\left({z}_{k}\right)}}}
\end{align}

\subsection{Computed Statistical Metrics}
\label{statmet}
Overall system performance has been compared using the statistical metrics: Accuracy, Precision, Recall and F1-score. These are calculated using the confusion matrix values such as true negative (TN), true positive (TP), false negative (FN), and false positive (FP). 

\vspace{3mm}

\noindent \textbf{\textit{Accuracy:}} Accuracy indicates the probability with which a trained model can predict correctly. It is defined as the proportion of correct guesses to the total number of predictions.
\begin{align}
Accuracy=\frac{TP+TN}{TP+TN+FP+FN}
\end{align}

\noindent \textbf{\textit{Precision:}} Precision signifies the chances of a predicted case being actually correct. It is defined as the ratio of true positives to total positives projected.
\begin{align}
Precision=\frac{TP}{TP+FP}
\end{align}

\noindent \textbf{\textit{Recall:}} Recall demonstrates the likelihood of the actual positive cases that can be predicted correctly. In other words, how many true positive conjectures have been produced out of all possible actually positive conjectures.
\begin{align}
Recall=\frac{TP}{TP+FN}
\end{align} 

\noindent \textbf{\textit{F1-Score:}} F1-Score is the harmonic mean of recall and precision. It is evaluated to assess how much a trained algorithm is error-free.
\begin{align}
F1\text{-}Score=2\times \frac{Precision\times Recall}{Precision+Recall}
\end{align}

\subsection{Working Principle of the Online Application}
\label{appwork}
As mentioned in section-\ref{Introduction}, the online application provides five COVID-19 diagnostic options along with the mortality risk prediction of the COVID-19 positive patients, as discussed below.

\vspace{3mm}

\begin{itemize} [noitemsep, topsep=0pt, leftmargin=*]
 \item [i.] \textit{\textbf{Symptoms:}} A list of possible symptoms is presented as checkboxes. The checked symptom by any individual will be marked as 1 and the unchecked symptom as 0. All the values will then be accumulated into an array of binary values.
  
\item [ii.] \textit{\textbf{Recorded Cough Audio:}}  This option allows the prediction of COVID-19 infection from the recorded audio of the cough sound of the individual who needs to be tested. It provides a start and stop recording button and the application requires the individual to voluntarily cough 5 times loudly. Upon selection of the predict button, the application will encode the cough sound to base64 string and send the encoded audio to a remote server which will then be reverted back to the original audio format and be processed for prediction.

\item [iii.] \textit{\textbf{Hematological parameters:}} A list of input parameters with their respective units are displayed on respective input fields, which will take numeric values as input.
 
\item [iv.] \textit{\textbf{Serum Raman Spectroscopy Image:}} Centrifuged Blood Serum Raman Spectral image obtained from any nearby pathological center should be uploaded to the application for further diagnosis. 
 
\item [v.] \textit{\textbf{ECG signal Image:}} ECG signal image obtained from any nearby pathological center using the device and report-paper configuration similar to Haider Khan et al. \cite{ecg_dataset} should be uploaded to the application for further diagnosis.

\item [vi.] \textit{\textbf{Mortality Risk Prediction:}} The COVID-19 positive patients will also be able to get their probable mortality risk due to the infection from two sets (7 and 9 features) of hematological parameter input obtained from routine blood test.

\end{itemize}

\vspace{3mm}

\noindent Once the input parameters of a particular option are completed, the predict button can be pressed. The application will then collect the data from input fields/checkboxes/encoded-audio and convert them to a JSON file, which will be sent to the pre-trained Machine Learning or Deep Learning model on a remote cloud server. The server continuously listens for a JSON file from the application. After receiving a JSON file, the developed model will predict the probability of COVID-19 infection and mortality risk, which will be sent back to the application and displayed in a popup dialogue box. This procedure in chronological order is illustrated by Figure-\ref{fig1} and \ref{fig7}.

\section{Result and Discussion}
\label{Result}

This research work explored a multimodal COVID-19 diagnostic procedure along with mortality risk prediction based on patient symptoms, recorded cough audio, hematological parameters, centrifuged serum Raman spectral image and ECG signal image. Both the machine learning and deep learning techniques have been investigated for these purposes along with a prototype online application implementation with the motivation of applying the concept of telepathology \cite{telepathology1, telepathology2} and telehealthcare into a reality in case of COVID-19 pandemic for providing point-of-care diagnostic services even in low-resource and remote areas.

\begin{figure}[ht]
\centering 
\resizebox*{12cm}{!}{\includegraphics{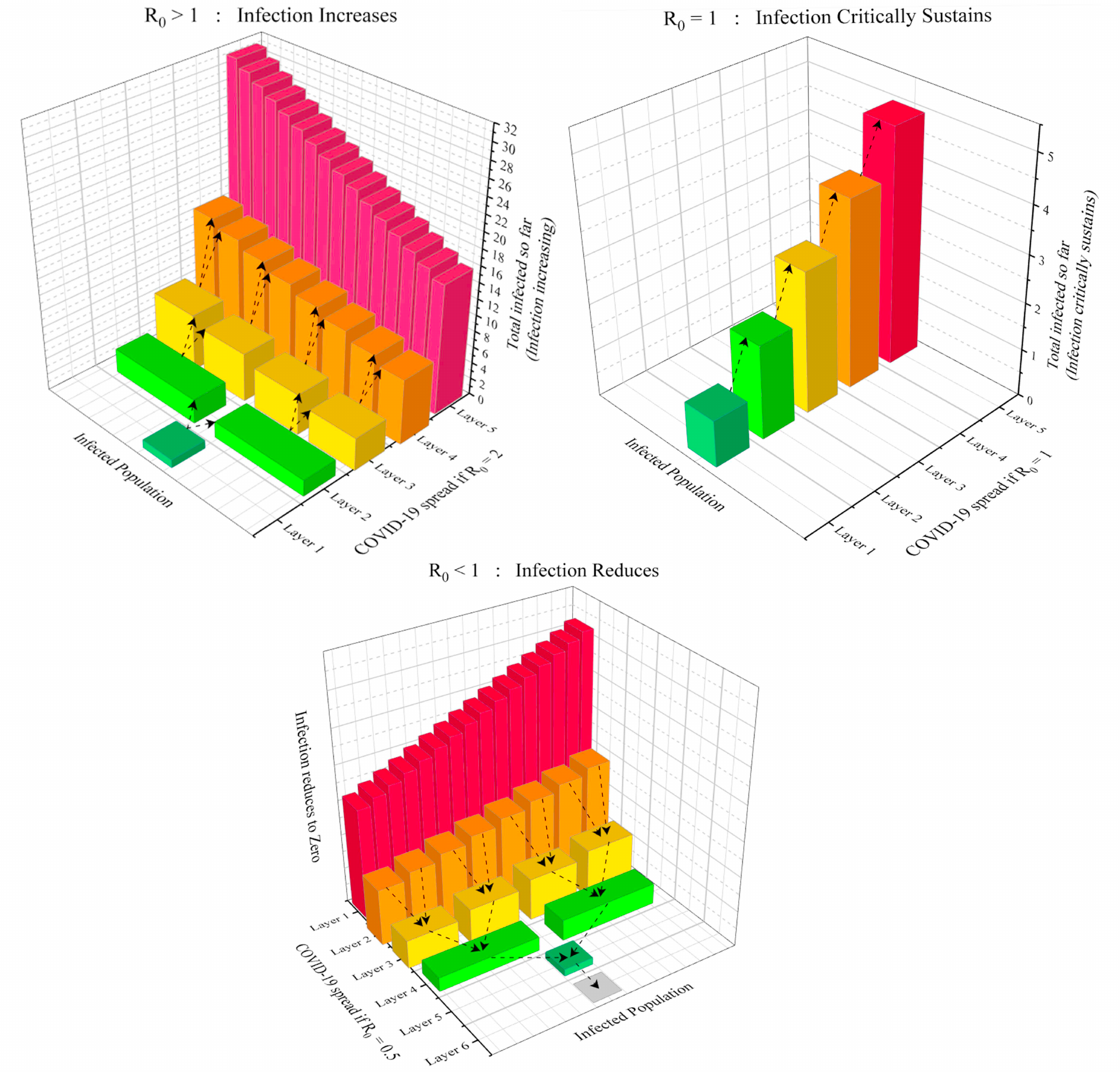}}
\caption{Effect of Basic Reproduction Number $(R_{0})$ on transmissibility/contagiousness of infectious disease.}
\label{fig6}
\end{figure}

At first, features from the possible symptoms of a patient are optimized and the five most important attributes are selected for prediction, as shown in Table-\ref{table4}. It can be observed from Figure-\ref{fig3}(a) that despite around 20\%-41\% positive correlation coefficients, none of the attributes have too strong significance in predicting COVID-19 infection undoubtedly. This is expected because these symptoms are not only specific to COVID-19 and may arise due to other diseases or physical complications. Nonetheless, a stacked ensemble machine learning model with RFC, XGBoost and SVM as the base-learners, and NB as the meta-learner achieved the highest overall accuracy of 77.59\%, as shown in Table-\ref{table5} (Exp-1), and could uniquely identify COVID-19 positive and negative patients with 74.06\% precision, 84.15\% recall and 78.78\% F1-Score. However, to make the prediction risk-free, nearly 100\% benchmark from a diagnostic solution is desired. A recall score of less than 100\% means that some COVID-19 negative patients would be falsely classified as positive, which upon quarantine would cause wastage of working hours. But, if few COVID-19 positive persons are falsely classified as negative due to sensitivity less than 100\% and they are not held in quarantine, then the misclassified persons will be enough to create a local epicenter of the virus outbreak. Thus, depending on the reproduction number $(R_{0})$, those misclassified patients roaming around freely may spread the virus through community transmission, as demonstrated by Figure-\ref{fig6} and the virus would persist in the community. \cite{blood1, reproduction_num}

A common symptom of SARS-CoV-2 infection is cough and it may initiate because of reflex action to irritants or inflammation within the lungs and windpipe. Such a pathway resembles a resonating structure with distinct fundamental frequency and vocal audio attributes. Inflammation due to COVID-19 may alter these attributes, which can be identified via machine learning audio analysis and can serve as the diagnostic method utilizing cough sounds. This study extracted seven audio parameters from the recorded cough sounds as listed in Table-\ref{table4} for COVID-19 diagnosis. Individually, the RFC algorithm showed the best performance with accuracy, precision, recall and F1-Score of 87.72\%, 85.25\%, 91.23\% and 88.14\% respectively. But with the proposed stacked ensemble machine learning model having RFC, XGBoost and DT as the base-learners and LR as the meta-learner, it can be observed that the ensemble model achieved a significant gain in accuracy, precision, recall and F1-Score to 95.65\%, 100\%, 92.86\% and 96.30\% respectively (Table-\ref{table5} Exp-2).

The emergence of multiple strains of the SARS-CoV-2 virus and their global dissemination throughout the population of diverse regions has manifested in both symptomatic and asymptomatic COVID-19 infected individuals. Diagnosing COVID-19 infection from symptoms or cough audio is nearly impossible in the case of asymptomatic patients because the majority of these symptoms have not evolved yet in them. Remarkably, both symptomatic and asymptomatic COVID-19 infected patients experience alterations in blood count, hematochemical profile, and the presence of antibody/antigen (IgM+, IgG+, IgM+/IgG++) in blood serum as a consequence of the immunological response against the viral infection. Hematological parameters from routine blood test showed the most promising result when a total of 25 parameters are used for COVID-19 diagnosis. In this case also, for an individual algorithm, the RFC demonstrated the highest accuracy, precision, recall and F1-Score of 97.12\%, 99.32\%, 94.81\% and 97.01\% respectively. Furthermore, the designed stacked ensemble machine learning model attained the performance metrics of 100\%, as shown in Table-\ref{table5} (Exp-3.1). This diagnostic method is a low-cost micro-invasive procedure and the sole expense factor is the number of parameters to be measured from the routine blood test. To reduce the cost, this study heuristically minimized the number of parameters to five, as listed in Table-\ref{table4} and Figure-\ref{fig3}(c). The low-cost solution comes at the expense of reduced accuracy of 95.24\% for the ensemble model (Table-\ref{table5}, Exp-3.2). Now with a reduced number of parameters, the model can correctly identify individuals with a recall of 90.32\%. Nonetheless, the model can successfully diagnose all the COVID-19 positive patients, which can be related to high correlation values of these parameters, seldom seen for other diagnostic approaches. Therefore, the five-parameter proposal can be a low-cost diagnostic approach.

The presence of dissolved chemicals in the blood serum that emerged as a result of an immune response to the COVID-19 viral infection can be detected using Raman spectroscopic analysis. Centrifuged blood serum Raman spectral data confirms the presence of antibody/antigen (IgG+, IgM+, IgG+/IgM+, etc.) through detection of abnormal concentration of specific dissolved chemicals (albumin, carotenoids, lipids, phospholipids, immunoglobulins, nucleic acids, tryptophan, etc.) by analyzing their molecular energy, especially, frequency characteristics. In experiment-4, this study investigated Raman spectral image dataset using different input image resolutions for the proposed custom CNN architecture to find out the best pixel dimension. The initial image resolution used is $32\times32$, then the successive higher resolutions are obtained by doubling the preceding resolution: $64\times64$, $128\times128$, $256\times256$, $512\times512$, and finally incremented to $800\times800$ so that the highest resolution commensurates with the original image resolution of the dataset. The comparative analysis of the performance metrics indicates that the $64\times64$ input resolution exhibits the best COVID-19 diagnostic result (Table-\ref{table5} Exp-4) with 100\% precision, and 99.80\% accuracy, recall and F1-Score. Below and above $64\times64$ resolution gradually worsens the diagnostic capabilities. Therefore, it can be concluded that the proposed Raman spectral image analysis model using the custom CNN architecture doesn’t require higher resolution but still achieves nearly the desired best performance. Moreover, the proposed deep learning computer vision solution in this study outperformed the machine learning model applied on the same dataset by Yin et al. \cite{raman5} and the discriminant analysis model applied on another serum Raman spectral dataset by Goulart et al. \cite{raman6}, as reviewed in section-\ref{2.2 Lit_rev}.

Alveolar sacs gradually start to become partially filled with fluid due to viral infection to the respiratory system, immunological alterations, and their cumulative effects. Continuous forced breathing caused by a lack of space owing to clogged alveolar volumes evokes cough, and the patient feels weary as a result of the extra effort required by lung and chest muscles to breathe. These factors may also impact the myocardium, resulting in ECG signal manifestations in COVID-19-infected patients. Experiment-5 of this study looked into these manifestations for diagnosing COVID-19 infected patients from their 12-lead ECG signal paper-based report images. In classifying persons with COVID-19 positive vs normal heartbeat, the VGG-16 computer vision transfer learning classifier algorithm achieved 99.55\% accuracy, 99.42\% precision, 100\% recall and 99.71\% F1-Score. Such a high classification result is noteworthy since the overall experiment (Exp-5) does not require the amalgamation of both manual and multi-software image pre-processing steps along with classification as compared to the contemporary studies discussed in section-\ref{2.2 Lit_rev} but achieved higher classification benchmark compared to those. As a result, the proposed classifier approach based on ECG signal image report in this article is significant for use in an online mobile application that can be utilized by users even with no computer-programming experience.

\begin{table*}[htp]
\caption{Performance metrics of the conducted experiments for COVID-19 diagnosis (Exp - 1 to 5) and Mortality Risk prediction (Exp - 6.1, 6.2).}
\centering
\begin{adjustbox}{width={\textwidth},totalheight={22cm},keepaspectratio}
{\begin{tabular}{cccccccc} \toprule

\multirow{2}{*}{{\bf{Exp-1:}} Symptoms} & \multicolumn{3}{c}{\bf{Base Learners}} & &  \bf{Meta Learner} & &  \multirow{2}{*}{\bf{Method}} \\ \cmidrule{2-4} \cmidrule{6-6}
  &  RFC  &  XGBoost  &  SVM  & &  NB  & & \\ \midrule
Accuracy  & 76.83\% & 76.83\% & 76.83\% & & \textbf{77.59\%} & &  Stacked \\
Precision   &  74.13\% & 74.13\% & 74.13\% & & 74.06\% & & Ensemble  \\
Recall   &  82.77\% & 82.77\% & 82.77\% & & \textbf{84.15\%} & & Machine  \\
F1-Score  &  78.21\% & 78.21\% & 78.21\% & & \textbf{78.78\%} & & Learning  \\

\midrule
\multirow{2}{*}{{\bf{Exp-2:}} Cough Audio} & \multicolumn{3}{c}{\bf{Base Learners}} & &  \bf{Meta Learner} & &  \multirow{2}{*}{\bf{{Method}}} \\ \cmidrule{2-4} \cmidrule{6-6}
 &  RFC  &  XGBoost  &  DT  & &  LR  & & \\ \midrule
Accuracy    &  87.72\% & 85.09\% & 83.33\% & & \textbf{95.65\%} & &  Stacked \\
Precision   &  85.25\% & 81.25\% & 80.65\% & & \textbf{100\%} & & Ensemble \\
Recall   &  91.23\% & 91.23\% & 87.72\% & & \textbf{92.86\%} & & Machine \\
F1-Score   &  88.14\% & 85.95\% & 84.03\% & & \textbf{96.30\%} & & Learning  \\
\midrule

\multirow{2}{*}{{\bf{Exp-3.1:}} Blood Test (25 features)}  & \multicolumn{3}{c}{\bf{Base Learners}} & &  \bf{Meta Learner} & &  \multirow{2}{*}{\bf{{Method}}} \\ \cmidrule{2-4} \cmidrule{6-6}
  &  RFC  &  XGBoost  &  SVM  & &  NB  & & \\ \midrule
Accuracy  & 97.12\% & 96.19\% & 92.95\% & & \textbf{100\%} & &  Stacked \\
Precision  & 99.32\% & 97.28\% & 96.48\% & & \textbf{100\%} & & Ensemble \\
Recall  & 94.81\% & 92.86\% & 88.96\% & & \textbf{100\%} & & Machine \\
F1-Score  & 97.01\% & 95.02\% & 92.57\% & & \textbf{100\%} & & Learning \\
\midrule

\multirow{2}{*}{{\bf{Exp-3.2:}} Blood Test (5 features)}  & \multicolumn{3}{c}{\bf{Base Learners}} & &  \bf{Meta Learner} & &  \multirow{2}{*}{\bf{{Method}}} \\ \cmidrule{2-4} \cmidrule{6-6}
 &   RFC  &  XGBoost  &  KNN  & &  NB  & & \\ \midrule
Accuracy  & 92.31\% & 88.46\% & 87.18\% & & \textbf{95.24\%} & &  Stacked \\
Precision  & 97.10\% & 88.82\% & 95.24\% & & \textbf{100\%} & & Ensemble \\
Recall &  87.01\% & 87.66\% & 77.92\% & & \textbf{90.32\%} & & Machine \\
F1-Score  & 91.78\% & 88.24\% & 85.71\% & & \textbf{94.92\%} & & Learning \\
\midrule

\multirow{2}{*}{{\bf{Exp-4:}} Raman Spectroscopy}  & \multicolumn{6}{c}{\bf{Image Resolution}} &  \multirow{2}{*}{\bf{{Method}}} \\ \cmidrule{2-7}

  &  $32\times32$  &  \textbf{$64\times64$}  &  $128\times128$  & $256\times256$ &  $512\times512$  & $800\times800$ & \\ \midrule
Accuracy  & 95.00\% & \textbf{99.80\%} & 96.67\% & 96.67\% & 95.00\% & 95.00\% & Custom \\
Precision  & 96.43\% & \textbf{100\%} & 96.55\% & 96.55\% & 96.43\% & 96.43\% & CNN \\
Recall  & 93.10\% & \textbf{99.80\%} & 96.55\% & 96.55\% & 93.10\% & 93.10\% & Architecture \\
F1-Score  & 94.74\% & \textbf{99.80\%} & 96.55\% & 96.55\% & 94.74\% & 94.74\% & \\
\midrule

{\bf{Exp-5:}} ECG Signal Image  & \multicolumn{6}{c}{\bf{Image Resolution}: $224\times224$} & {\bf{{Method}}} \\ \midrule
Accuracy  & \multicolumn{6}{c}{99.55\%} & Transfer \\
Precision  & \multicolumn{6}{c}{99.42\%} & Learning \\
Recall  & \multicolumn{6}{c}{100\%} & (VGG-16) \\
F1-Score  & \multicolumn{6}{c}{99.71\%} & \\
\midrule

\multirow{2}{*}{{\bf{Exp-6.1:}} Blood Test (7 features)}  & \multicolumn{3}{c}{\bf{Base Learners}} & &  \bf{Meta Learner} & &  \multirow{2}{*}{\bf{{Method}}} \\ \cmidrule{2-4} \cmidrule{6-6}
  &  RFC  &  XGBoost  &  SVM  & &  RFC  & & \\ \midrule
Accuracy  & 94.17\% & 90.00\% & 89.17\% & & \textbf{95.83\%} & &  Stacked \\
Precision  & 95.14\% & 90.60\% & 89.57\% & & 93.33\% & & Ensemble \\
Recall  & 93.64\% & 89.51\% & 88.74\% & & \textbf{100\%} & & Machine \\
F1-Score  & 94.05\% & 89.82\% & 88.99\% & & \textbf{96.55\%} & & Learning \\
\midrule

\multirow{2}{*}{{\bf{Exp-6.2:}} Blood Test (9 features)}  & \multicolumn{3}{c}{\bf{Base Learners}} & &  \bf{Meta Learner} & &  \multirow{2}{*}{\bf{{Method}}} \\ \cmidrule{2-4} \cmidrule{6-6}
 &   RFC  &  XGBoost  &  SVM  & &  RFC  & & \\ \midrule
Accuracy  & 80.00\% & 80.00\% & 72.00\% & & \textbf{92.00\%} & &  Stacked \\
Precision  & 72.73\% & 72.73\% & 61.54\% & & \textbf{92.86\%} & & Ensemble \\
Recall &  80.00\% & 80.00\% & 80.00\% & & \textbf{92.86\%} & & Machine \\
F1-Score  & 76.19\% & 76.19\% & 69.57\% & & \textbf{92.86\%} & & Learning \\

\bottomrule
\end{tabular}}
\end{adjustbox}
\label{table5}
\end{table*}

Hospitals and healthcare facilities have reported a dearth of hospital and ICU beds, as well as a lack of manpower and logistic support, in order to handle the massive influx of COVID-19-infected patients with equal care during the peak seasons. COVID-19 infected patients with severe comorbidities, such as cancer, experience a far worse health situation until they receive ICU care, which they acquire after overcoming the barrier of overloaded hospital COVID wards and overworked hospital staff. COVID-19 diagnostic report combined with the Mortality Risk projection can ease up the scenario. The hospital staff and caregivers can manage the available resources with the highest possible efficiency if they are informed regularly about the percentage probability of mortality risk of a COVID-19 positive patient who either needs to be admitted to the hospital cohort or needs mechanical ventilation with critical care support who has already been admitted to the general COVID ward. To resolve this concern, Exp-6.1 and 6.2 of this study focused on predicting mortality risk using optimized routine blood test parameters (Table-\ref{table4} and Figure-\ref{fig3}(d, e)) obtained from two different COVID-19 positive patient dataset. A similar stacked ensemble machine learning model has been proposed for both the experiments incorporating RFC, XGBoost, SVM classifiers as the base-learner and RFC algorithm as the meta-learner. The Exp-6.1 considering 7 attributes achieved 95.83\% accuracy, 93.33\% precision, 100\% recall and 96.55\% F1-Score; while the Exp-6.2 considering 9 attributes achieved an accuracy score of 92\% and precision, recall and F1-Score of 92.86\% each, as shown in Table-\ref{table5}. Both the experiments considering only optimized 7 and 9 blood-test parameters outperformed the contemporary research works on the same datasets conducted by Hoon Ko et al. \cite{mortality2} and Mahdavi et al. \cite{mortality3} who considered 28 and 37 both non-invasive and micro-invasive type of features respectively for mortality risk prediction, as outlined in section-\ref{2.2 Lit_rev}.

The Basic Reproduction number \cite{blood1, reproduction_num}, designated $R_{0}$, is one of the quantifying factors of the dissemination capability of an infectious disease. The parameter gives an estimation of the number of secondary infections caused by an already infected individual provided that everyone in the community is vulnerable, as illustrated in Figure-\ref{fig6}. Therefore, two large-scale management approaches are available to reduce the spread of the COVID-19 virus. The first approach is to control the already infected individuals through mass testing followed by mandatory quarantine/isolation with proper treatment, while the second option is to control the susceptible population through mass vaccination. The implication of mass testing requires fast, point-of-care and widely available diagnostic procedures that are also reliable in terms of classification, i.e. the overall procedure reduces the false positive (FP) or false negative (FN) misclassifications to a bare minimum so that the misclassified FN individuals do not contribute increment of the Basic Reproduction number and the misclassified FP individuals do not lose valuable working hours. All the proposed multimodal COVID-19 diagnostic approaches except the Symptoms experiment in this article have exhibited a benchmark performance of more than 90\%, few of them even achieved 100\% precision/recall values. 

This study needed to be conducted in the form of separate experiments due to the unavailability of any public dataset of COVID-19 positive/negative individuals containing data regarding all the proposed diagnostic modes of a particular person such as symptoms, cough audio, hematological parameters, serum Raman spectroscopy and ECG signal image. Even so, the authors of this article strongly believe that the diagnostic results will be more reliable if predictions from all the modes of data collected from a single person are fused together to give the final classified output through the fusion of multiple machine and deep learning models together. The blood tests, Raman spectrophotometer and ECG scanner are widely available in diagnostic centers even in the underdeveloped countries and these tests require little time to produce the report. As a result, the solution proposed in this article can be used as an emergency substitute for state-of-the-art gold standard COVID-19 diagnostic tests, proving it to be an ideal alternative diagnostic tool for low-income as well as remote areas and facilitating on-site testing due to its convenient cloud-based classification via mobile application.

\begin{figure}[ht]
\centering 
\resizebox*{15cm}{!}{\includegraphics{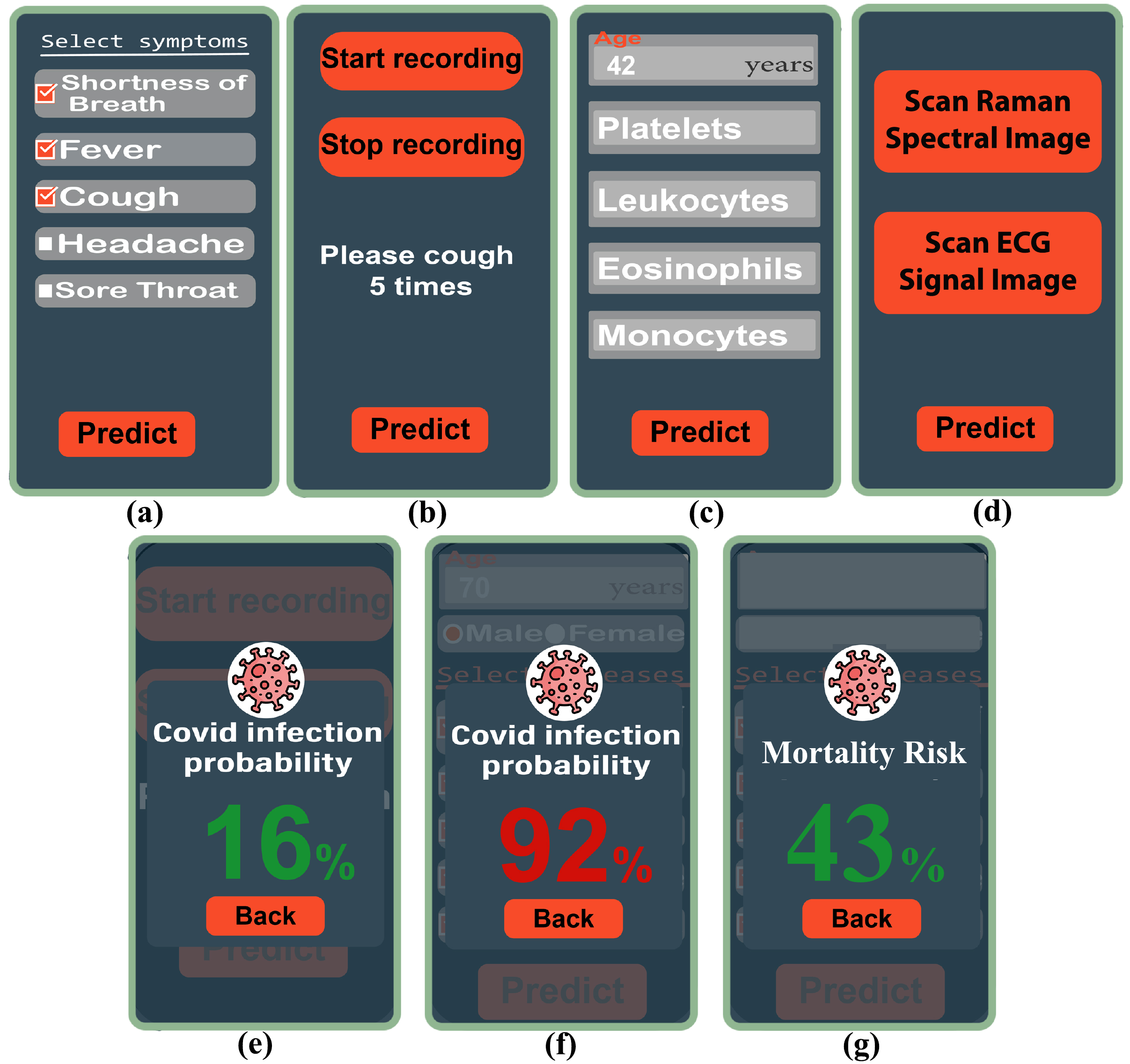}}
\caption{Online application user input and output interface prototype.}
\label{fig7}
\end{figure}

Figure-\ref{fig7}(a-d) illustrates the user input interface of the prototype online application for symptoms, cough audio, blood test parameters (5 features), Raman spectral image, ECG signal report image for COVID-19 diagnosis, and blood test parameters (7 features and 9 features) for mortality risk prediction respectively. The classification results and mortality risk predictions are demonstrated through a separate dialog box, as portrayed in Figure-\ref{fig7}(e-g).

\clearpage

\section{Conclusion}
\label{Conclusion}
This paper proposes and experimentally demonstrates a novel online framework to identify positive COVID-19 patients along with their mortality risk prediction using machine learning and computer vision. It can be accessed through a cloud based smartphone application by both non-ambulatory and hospitalized patients. The application provides five modes of diagnosis using symptoms, cough sound, hematological biomarkers, Raman spectral data from blood specimen, and ECG signal image with an accuracy of 77.59\%, 95.65\%, 95.24\% (for 5 blood features, but 100\% for 25 blood features), 99.80\%, and 99.55\% respectively. The algorithms exhibited a sensitivity of 100\% for the features obtained from blood and sound, which indicates correct identification of COVID-19 positive patients. The multimodal diagnosis helps to better classify possible infectees. Also, the attributes required for the predictions are optimized and minimized. This is imperative to reduce the cost of COVID-19 detection, especially from routine blood test, which is available in most of the clinical and rural medical infrastructure in even underdeveloped and developing countries to provide an alternative feasible low-cost method of diagnosis, when there is scarcity of COVID-19 PCR test kits. The online, scalable and real-time essence of the diagnostic platform can provide an additional fast and immediate screening method of the continuous mutating virus and control the future potential waves of infection. Moreover, the app will help alerting the respective healthcare personnel by providing patient mortality risk prediction so that they can best utilize their limited critical care resources for the patients who are in dire need. The authors hope that the proposed technique herein demonstrates the power of telehealthcare as an easy, widespread, low-cost, and scalable diagnostic solution for COVID-19 as well as future pandemics.

\section{Supplementary Materials}
\label{sup_mat}
\begin{itemize}
    \item[(a)] Online Application Link: [Link will be added to the final accepted version of the article]
    \item[(b)] Proof of dissimilarities of the ECG images:
    
    \url{https://github.com/MDMohiUddinKhan/Dissimilar-ECG}
\end{itemize}

\section{Author Statements}
\noindent\textit{Abdullah Bin Shams:} Idea generation, Writing - Quality control, Project administration and Supervision

\noindent\textit{Md. Mohsin Sarker Raihan:} Idea generation, Exp-2, Exp-3, Exp-4, Exp-5, Supervision

\noindent\textit{Md. Mohi Uddin Khan:} Exp-2, Exp-5, Literature review, Graphics, Writing - Original Draft and Editing

\noindent\textit{Ocean Monjur:} Exp-1, Exp-6, Literature review

\noindent\textit{Rahat Bin Preo:} Theme picture and app console design, Heroku application development

\vspace{5pt}

\noindent Finally all authors reviewed and discussed on the article, provided critical feedback and contributed to the final manuscript.

\section{Research Funding} 
This research did not receive any specific grant from funding agencies in the public, commercial, or not-for-profit
sectors. 

\section{Declaration of Competing Interest} The authors declare that they have no known competing financial interests or personal relationships that could
have appeared to influence the work reported in this article.

\clearpage
\bibliographystyle{elsarticle-num} 
\bibliography{references}

\end{document}